\documentclass{article}

\usepackage[final]{corl_2026} 
\usepackage{amsmath,amssymb,amsfonts}
\usepackage{graphicx}
\usepackage{bbm}

\usepackage{booktabs}
\usepackage{algorithm}
\usepackage{algpseudocode} 

\usepackage{array}
\usepackage{url}
\usepackage{xcolor}
\usepackage{booktabs}
\usepackage{makecell}
\usepackage{tabularx}
\usepackage[table]{xcolor}
\usepackage{pgf}
\usepackage{wrapfig}
\usepackage{capt-of}

\newcommand{\methodshort}{\textsc{B2FF}}

\title{Back to the Familiar Future: Failure Recovery for VLA Policies via Pre-Imagined Milestone Selection}

\author{
\begin{tabular}{c}
\textbf{Suyeon Shin}$^{1}$ \quad
\textbf{Juwon Kim}$^{2}$ \quad
\textbf{Hyeonbin Park}$^{3}$ \quad
\textbf{Hyunseo Kim}$^{1}$ \\
\textbf{Hyundo Lee}$^{1}$ \quad
\textbf{Hyung-Sin Kim}$^{1, \dag}$ \quad
\textbf{Byoung-Tak Zhang}$^{1,4 \dag}$ \\[0.4em]
{\normalfont\small
$^{1}$Seoul National University \quad
$^{2}$Yonsei University \quad
$^{3}$Soongsil University \quad
$^{4}$Tommoro Robotics} \\[0.25em]
{\normalfont\ttfamily\small
\{syshin, btzhang\}@bi.snu.ac.kr} \\
\end{tabular}
}

\begin{document}

\maketitle

\begingroup
\renewcommand{\thefootnote}{\dag}
\footnotetext{Corresponding authors.}
\endgroup


\begin{abstract}
Vision-language-action (VLA) policies can deviate from nominal trajectories during manipulation, even when tasks remain physically feasible. Recovering from these deviations is challenging, as they push the policy into unfamiliar state spaces where direct re-planning frequently destabilizes action sequences. We propose Back to the Familiar Future (B2FF), a recovery framework for foresight-driven VLAs that leverages future visual conditioning as a recovery interface. Before execution, the VLA generates a milestone bank of familiar future states conditioned on the clean initial observation. At recovery time, a recoverability-aware selector selects a recovery milestone from this bank and enforces it as a fixed visual goal. This enables the VLA to robustly map off-trajectory observations back to a familiar future. On failure-injected LIBERO, under controlled recovery timing aligned with the injected failure, B2FF increases the average success rate of a baseline VLA from 56.3\% to 74.0\%, demonstrating that pre-imagined milestones can guide recovery without fine-tuning the low-level action generator. Project page: https://syshin-snu.github.io/b2ff/.

\end{abstract}

\begin{center}
\includegraphics[width=0.97\linewidth]{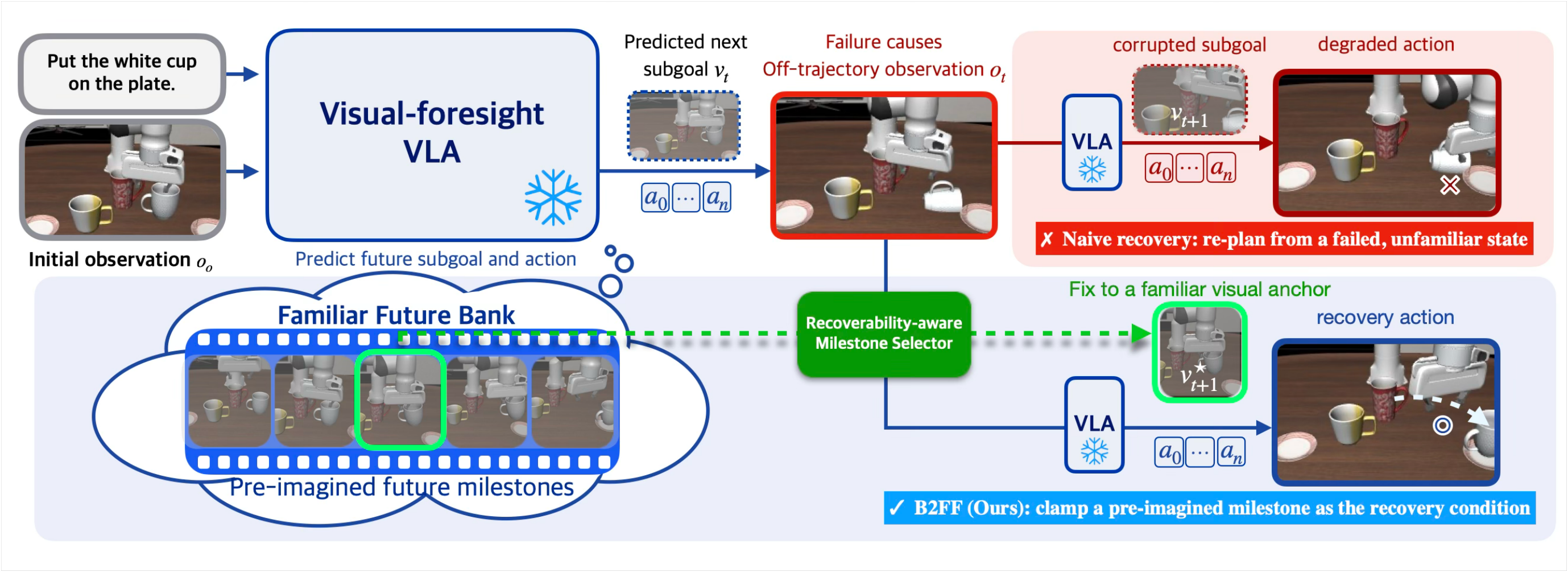}
\captionof{figure}{\textbf{Overview of BACK TO THE FAMILIAR FUTURE}.
Instead of re-predicting a future from an off-trajectory observation, B2FF selects a pre-imagined milestone from the familiar future bank and uses it as the fixed visual anchor for action-only denoising.}
\label{fig:overview}
\end{center}

\section{Introduction}

Vision-language-action (VLA) policies have advanced language-conditioned robotic manipulation by connecting vision-language representations with robot control \cite{brohan2022rt,o2024open,rt2,octo,openvla,pizero}. However, strong nominal performance does not guarantee reliable recovery once execution deviates, and recent work has identified failure recovery as key challenge for deploying robotic manipulation policies \cite{fiper2025,safe2025,failsafe2025,dai2025racer}. During manipulation, a learned policy can drift away from demonstrated trajectories due to compounding action errors, contact dynamics, or unexpected state changes. 
The task may remain physically feasible, but the policy must now act from off-trajectory observations that are less familiar than those encountered during nominal execution. 
Recovery is therefore challenging not because the required low-level skills are absent, but because directly querying the policy from these unfamiliar states can produce unreliable actions.


Existing recovery and correction methods tackle failure recovery by synthesizing failure data, generating corrective actions, adding reflection modules, or modifying the execution pipeline~\cite{failsafe2025,phoenix2025,dream2fix2026,dcdp2026}. 
Because these approaches predominantly intervene at the action-generation level, they often require updating or bypassing the core policy.
Foresight-driven VLAs, however, present a distinct intervention point: their action generation is inherently mediated by an intermediate future representation. 
We therefore ask whether recovery can be achieved by providing a frozen VLA with a more suitable visual guide after failure, rather than modifying the action generator.

We propose \textsc{Back to the Familiar Future} (B2FF), an inference-time recovery framework for pre-trained foresight-driven VLA policies. B2FF operates under entirely frozen VLA, permitting no weight updates or fine-tuning. Under this strict constraint, B2FF treats recovery as selecting an action-guiding visual condition rather than the modification of action outputs. 
Before execution, B2FF constructs a \textit{familiar future bank} by recursively querying the frozen VLA from the clean initial observation and storing the generated future-image subgoals as milestones.
At recovery time, a recoverability-aware selector chooses the optimal milestone from this bank and enforces it as a future visual condition.
The frozen VLA then generates actions from the actual off-trajectory observation toward this familiar anchor. Figure~\ref{fig:overview} illustrates the high-level recovery mechanism.
We evaluate B2FF on standard and failure-injected LIBERO \cite{libero2023}, covering nominal execution, policy-induced failures, and injected disruptions. Furthermore, real-world experiments demonstrate that these selected milestones induce highly interpretable recovery behaviors.

\section{Related Work}

\paragraph{Foresight-driven VLA policies.}
Recent robot policies increasingly use generated future states as intermediate structures for action generation, including video plans, future images, visual chain-of-thought, motion forecasts, and unified future-action token generation \cite{du2024vlp,du2023unipi,cotvla2025,dreamvla2025,flowvla2025,udvla2026}. These methods show that imagined visual states can support nominal manipulation by providing explicit targets or reasoning variables. B2FF builds on this foresight paradigm but studies a different problem: recovery after off-trajectory deviations. Rather than generating a new future from the failed observation, B2FF selects a pre-imagined familiar future as the recovery condition.

\paragraph{Failure recovery for VLAs.}
Recent recovery methods for robot policies typically intervene by producing corrective behavior after failure. 
Some approaches synthesize failure-recovery data or train recovery policies from counterfactual failures and language-guided correction examples \cite{failsafe2025,dream2fix2026,dai2025racer}. 
Others use reflection modules, VLM prompting, or dynamic action correction to revise the robot's next actions during execution \cite{phoenix2025,chen2024automating,dcdp2026}. 
While these methods primarily view recovery as learning, reasoning about, or directly generating corrective actions, B2FF proposes an alternative paradigm. By strictly maintaining the low-level action generator unchanged, B2FF formulates recovery as a visual conditioning problem: selecting which pre-imagined milestone should condition action generation to guide the frozen VLA from the current off-trajectory observation.

\paragraph{Progress markers and recovery conditions.}
Progress-aware methods structure long-horizon manipulation by generating visual subgoals from task-progress knowledge~\cite{taksie2025}, predicting intermediate subgoals through backward planning~\cite{lbp2025}, guiding diffusion policies with progress signals~\cite{progressvla2026}, or rewinding to progress-aware milestones when task progress stalls~\cite{spr2026}. 
These methods use subgoals or milestones to organize nominal task progress and decide where execution should proceed or rewind. 
In contrast, B2FF employs a pre-imagined milestone sequence not for tracking nominal progress, but as candidate visual guides for recovery: after failure, it selects which milestone should condition action generation from the current off-trajectory observation.

\section{Method}
\begin{figure}[t]
\centering
\includegraphics[width=\linewidth]{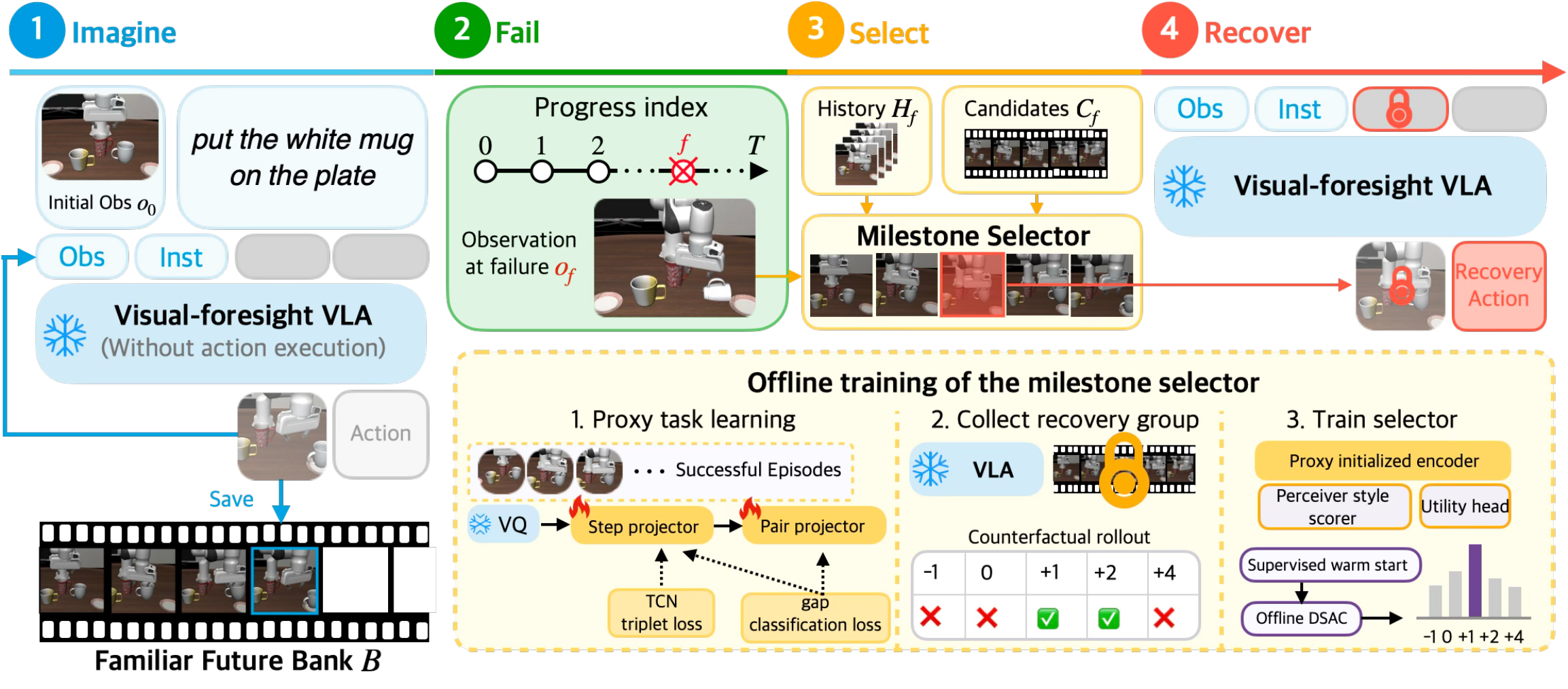}

\caption{
\textbf{B2FF pipeline.}
Before execution, the frozen VLA builds a familiar future bank.
At recovery time, B2FF constructs local candidates, selects a recoverable milestone, and uses it as the fixed future-image anchor.
The selector is trained offline from counterfactual recovery labels.
}
\label{fig:b2ff-pipeline}
\end{figure}

We propose \textsc{Back to the Familiar Future} (B2FF), an inference-time recovery framework for foresight-driven VLA policies. Following an off-trajectory deviation, B2FF avoids re-imagining the future directly from the current unfamiliar state. Instead it selects a visual milestone predicted prior to execution and uses it as a fixed condition for recovery. The framework consists of three steps: 
imagining a familiar future bank before execution (Sec.~\ref{sec:familiar-future-bank}), entering an off-trajectory recovery context (Sec.~\ref{sec:detect}), selecting the most recoverable familiar future from the bank, and fixing the selected familiar future as the visual condition for action-only denoising (Sec.~\ref{sec:selection}). Figure~\ref{fig:b2ff-pipeline} summarizes the online recovery pipeline and offline selector training.

\paragraph{Visual Foresight as a Recovery Interface.}

We assume access to a frozen foresight-driven VLA policy \(\pi_\theta\) with an explicit visual future interface. 
Let \(I\) be the language instruction, \(o_t\) the current image observation, \(v_t\) a future-image subgoal, and \(\mathbf a_t=a_{t:t+L-1}\) a length-\(L\) action chunk. 
We operate the VLA policy in two distinct inference modes:
\[
\underbrace{\pi_\theta(v_t,\mathbf a_t\mid I,o_t)}_{\text{joint subgoal--action generation}},
\qquad
\underbrace{\pi_\theta(\mathbf a_t\mid I,o_t;\, v_t\leftarrow v^\star)}_{\text{action-only denoising with a fixed image subgoal}} .
\tag{1}
\]
The first mode corresponds to standard joint generation, where the frozen VLA predicts both subgoal and actions from the current observation. 
The second mode performs action-only denoising by setting the future-image tokens to a fixed \(v^\star\). 
In both settings, the future visual condition guides action generation. During nominal execution, we employ the joint generation mode; conversely, during recovery, once a specific \(v^\star\) is selected, we switch to the action-only denoising mode.

\subsection{Familiar Future Bank}
\label{sec:familiar-future-bank}

A familiar future is a visual milestone imagined before any execution-induced visual shift is observed. 
At the beginning of each episode, B2FF constructs an imagination bank \(\mathcal B=\{\tilde{v}_1,\tilde{v}_2,\ldots,\tilde{v}_M\}\) by recursively querying the future-image marginal of the frozen VLA. 
Initializing with \(\tilde{v}_0=o_0\), each imagined milestone is generated as \(\tilde{v}_m\sim\pi_\theta(v\mid I,\tilde{v}_{m-1})\) for \(m=1,\ldots,M\).  
Actions are not executed during this imagination rollout.
Since the bank is generated from the initial observation, it provides familiar visual anchors rather than futures predicted from a failed state.

\subsection{Recovery-Mode Entry and Candidate Set}
\label{sec:detect}
Recovery begins at an index \(f\), which specifies when B2FF switches from nominal to recovery mode.
In controlled failure-injected evaluation, \(f\) is aligned with the injected perturbation; in the online-triggered variant, it is estimated from proprioceptive history.
\[
f =  \min\{t \mid D(p_{1:t}) > \tau_D\}.
\tag{2}
\]
Here, \(p_{1:t}\) denotes the proprioceptive history up to time \(t\), \(D\) is the learned detector, and \(\tau_D\) is the detection threshold. 
If the threshold is not crossed, B2FF remains in nominal execution mode. We treat trigger estimation as an interchangeable entry point.

Once recovery mode is entered, B2FF uses the current off-trajectory observation \(o_f\) and recent observation history \(H_f\) to build a local candidate set from the familiar future bank. 
Given a fixed offset set \(\Delta=\{\delta_1,\ldots,\delta_K\}\), the candidate set is:
\[
C_f=\{\tilde v_{f+\delta} \mid \delta\in\Delta,\ 1\le f+\delta\le M\}\subset\mathcal B.
\tag{3}
\]

\subsection{Recoverability-Aware Milestone Selection}
\label{sec:selection}
Once entering recovery mode at index \(f\), recoverability-aware selector \(F_\phi\) selects a familiar future \(v^\star\) from the local candidate set \(C_f\) by scoring each candidate \(\tilde v \in C_f\) under the recovery context \((o_f, H_f, C_f)\):
\[
v^\star = \arg\max_{\tilde v\in C_f} F_\phi(\tilde v \mid o_f,H_f,C_f).
\tag{4}
\]

Here, \(F_\phi(\tilde v\mid o_f,H_f,C_f)\) represents the score for using \(\tilde v\) as the fixed future-image subgoal from the current failure-state observation. 
To compute this score, the selector first encodes \(o_f\), \(H_f\), and \(C_f\) using the frozen visual tokenizer. The extracted features are then processed by lightweight learned projectors, a Perceiver-style attention module~\cite{jaegle2021perceiver} that summarizes the candidate-context features, and finally an MLP scoring head that outputs one scalar score per candidate.
We then use \(v^\star\) as the fixed future-image subgoal in the action-only denoising mode of Eq.~(1).

\paragraph{Selector training.}
The selector is trained offline in three stages. 
First, proxy temporal initialization provides a progress-aware initialization for subsequent recoverability learning using successful trajectories.
Following time-contrastive representation learning~\cite{sermanet2018time}, we train a step projector for individual observations and future-image milestones, and a pair projector for observation--future pairs:
\(
\mathcal L_{\mathrm{proxy}}
=
\mathcal L_{\mathrm{TCN}}
+
\beta_{\mathrm{gap}}\mathcal L_{\mathrm{gap}}.
\)
\(\mathcal L_{\mathrm{TCN}}\) is a triplet-style time-contrastive loss that pulls VLA-predicted future images toward temporally nearby trajectory frames and pushes them away from distant frames, while \(\mathcal L_{\mathrm{gap}}\) is a cross-entropy loss over discretized temporal-gap bins predicted from the pair representation. 
After pretraining, we discard the auxiliary gap head and retain the step and pair projectors for selector feature construction.

To enable subsequent training stages, we collect offline counterfactual rollout data for selector training by applying the action-only denoising mode of Eq.~(1) to each candidate milestone. 
For each injected training failure and each candidate \(c\in C_f\), we set \(v^\star=c\) and roll out the frozen VLA from the same failure-state observation, producing a binary label \(y_c\in\{0,1\}\) that indicates final task success.
These counterfactual rollouts provide offline candidate-wise supervision for the selector.

Second, the supervised warm-start trains the selector from counterfactual labels before RL-style fine-tuning. 
We use two complementary objectives. 
The BCE term provides candidate-wise recoverability supervision. 
Meanwhile, the grouped term encourages successful candidates to receive higher normalized scores than failed ones within the same failure context, without choosing an arbitrary single positive candidate when multiple milestones succeed.
Using the selector score \(s_c\) for each candidate \(c\in C_f\), we use a candidate-wise BCE loss:
\[
\mathcal L_{\mathrm{BCE}}
=
-\frac{1}{|C_f|}
\sum_{c\in C_f}
\left[
y_c\log\sigma(s_c)
+
(1-y_c)\log(1-\sigma(s_c))
\right].
\tag{5}
\]
When \(Y_f>0\), we additionally apply a grouped ranking loss:
\[
\mathcal L_{\mathrm{group}}
=
-\frac{1}{Y_f}
\sum_{c\in C_f}
y_c s_c
+
\log
\sum_{c\in C_f}
\exp(s_c),
\qquad
Y_f=\sum_{c\in C_f}y_c.
\tag{6}
\]
For all-fail groups, we omit \(\mathcal L_{\mathrm{group}}\). 
We optimize the weighted supervised objective \(\mathcal L_{\mathrm{sup}}=\lambda_{\mathrm{BCE}}\mathcal L_{\mathrm{BCE}}+\lambda_{\mathrm{group}}\mathcal L_{\mathrm{group}}\).

Finally, we refine the selector with a one-step discrete actor-critic-style objective inspired by maximum-entropy actor-critic methods~\cite{haarnoja2018soft,christodoulou2019soft}. 
For each failure context \((o_f,H_f,C_f)\), each candidate \(c\in C_f\) is treated as a discrete recovery choice, and the recorded success label \(y_c\) serves as its utility target. 
Since this is a one-step candidate choice, the critic directly regresses to the recorded success label. 
We fit two critic heads using Huber regression~\cite{huber1992robust}:
\[
\mathcal L_Q
=
\mathbb E
\left[
\frac{w_f}{|C_f|}
\sum_{c\in C_f}
\sum_{k=1}^{2}
\operatorname{Huber}
\left(
Q_k(c\mid o_f,H_f,C_f)-y_c
\right)
\right].
\tag{7}
\]
The actor-style term then shifts the selector toward candidates with high predicted recovery utility:
\[
\mathcal L_{\mathrm{actor}}
=
\mathbb E
\left[
w_f
\sum_{c\in C_f}
\operatorname{softmax}(\mathbf{s})_c
\left(
\alpha_{\mathrm{low}}
\log\operatorname{softmax}(\mathbf{s})_c
-
\min_{k\in\{1,2\}}Q_k(c\mid o_f,H_f,C_f)
\right)
\right],
\tag{8}
\]
where \(\mathbf{s}\) denotes the selector scores over \(C_f\), \(w_f\) is the group weight, and \(\alpha_{\mathrm{low}}\) is the entropy coefficient used for the lower-entropy actor update. 
We optimize the final fine-tuning objective \(\mathcal L_{\mathrm{ft}}=\mathcal L_Q+\mathcal L_{\mathrm{actor}}+\lambda_{\mathrm{BC}}\mathcal L_{\mathrm{group}}\), where the last term reuses the grouped positive-candidate objective as a behavior-cloning regularizer toward successful recovery choices. 
This fine-tuning stage biases the selector toward milestones with higher recorded recovery success, while keeping the frozen VLA as the sole low-level action generator.


\providecommand{\methodname}{B2FF}
\providecommand{\tbd}{\textsc{TBD}}
\newcommand{\repro}{\ensuremath{^{\dagger}}}

\definecolor{HeatBlue}{HTML}{5B9BD5}
\definecolor{B2FFBlue}{HTML}{1F4E9D}
\definecolor{B2FFBg}{HTML}{EEF5FF}
\definecolor{OracleBg}{HTML}{F1F1F1}
\definecolor{OracleText}{HTML}{222222}
\definecolor{TBDGray}{HTML}{8A8A8A}

\newcommand{\cmark}{\ensuremath{\checkmark}}
\newcommand{\na}{--}

\newcommand{\heat}[1]{%
  \pgfmathtruncatemacro{\heatshade}{min(55,max(8,round((#1-20)*0.65)))}%
  \begingroup
  \edef\x{\endgroup\noexpand\cellcolor{HeatBlue!\heatshade!white}}%
  \x #1%
}

\newcommand{\bhe}[1]{%
  \pgfmathtruncatemacro{\heatshade}{min(48, max(6, round((#1-20)*0.55)))}%
  \begingroup
  \edef\x{\endgroup\noexpand\cellcolor{HeatBlue!\heatshade!white}}%
  \x \textbf{#1}%
}

\newcommand{\ocell}[1]{%
  \cellcolor{OracleBg}\textcolor{OracleText}{#1}%
}

\newcommand{\bocell}[1]{%
  \cellcolor{OracleBg}\textcolor{OracleText}{\textbf{#1}}%
}

\newcommand{\tbdcell}{%
  \cellcolor{OracleBg}\textcolor{TBDGray}{\emph{TBD}}%
}

\section{Experiments}
\label{sec:experiments}

We evaluate \methodname{} as a recovery interface for frozen foresight-driven VLAs. 
For B2FF-specific ablations, the frozen VLA, familiar future bank, and recovery interface are fixed; variants differ only in the recovery-time future visual condition or selector design. 
Specifically, we ask: (Q1) Does conditioning on a pre-imagined familiar future improve recovery over failed-state re-planning? 
(Q2) Does recoverability-aware selection outperform fixed or visually nearest milestones? 
(Q3) Which selector-training and input choices drive the gains?

\subsection{Experimental Setup}
\label{subsec:exp-setup}

\paragraph{Benchmarks.}
We evaluate on standard LIBERO rollouts and a failure-injected LIBERO variant. 
The former evaluates recovery on naturally occurring policy failures while verifying that successful rollouts remain unaffected by intervention. 
The latter randomly injects one execution failure, enabling controlled comparison across gripper shifts, object shifts, and object laydowns.
These failures cover end-effector misalignment, object displacement, and unexpected object pose or orientation. We explicitly exclude irreversible cases such as workspace exits, severe occlusions, physical damage, or unachievable instructions.

\paragraph{Failure injection timing.}
Perturbations are concentrated around critical
manipulation phases, particularly near grasping.
As shown in Figure~\ref{fig:recovery_diagnostics}(a),
the injection times are centered near the grasp
event, with a slight pre-grasp bias
(mean timing gap of -3.3 timesteps).
This protocol targets deviations that can disrupt
grasping and subsequent manipulation while
preserving task feasibility.

\paragraph{Recovery protocol.}
We use a deterministic familiar future bank with \(M=12\) and candidate offsets \(\Delta=\{-1,0,+1,+2,+4\}\), covering rollback, retry, and skip-ahead hypotheses. 
In controlled experiments, recovery timing is aligned with the injected perturbation to isolate milestone selection; the online triggered variant estimates recovery timing from proprioceptive history. 
The same trigger-selection procedure can be re-applied if later recovery triggers occur, although our controlled evaluation injects one perturbation per episode. 
The visual-nearest baseline selects the candidate with the highest cosine similarity to the failure observation in the frozen visual-tokenizer feature space. 
Candidate-wise five-chunk rollouts are used only for selector training and candidate-selection upper-bound analysis; test-time recovery performs a single selector forward pass without trial rollouts.

\subsection{Quantitative Results on Failure-Injected and Standard LIBERO}
\label{subsec:main-results}

Table~\ref{tab:main_summary} summarizes the standard and failure-injected LIBERO results. 
We compare improvements within distinct policy families: DCDP~\cite{dcdp2026} adds dynamic action correction to DP~\cite{diffusionpolicy2023}, SPR-VLA~\cite{spr2026} adds progress-aware planning and rewind to MolmoAct~\cite{lee2025molmoact}, and B2FF builds on UD-VLA~\cite{udvla2026} by changing the recovery-time visual condition of the frozen policy.

On the controlled failure-injected benchmark, B2FF improves UD-VLA from 56.3\% to 74.0\% average success, a +17.7 percentage-point gain. 
This improvement is substantially larger than the gains observed in other baseline families, such as DCDP over DP (+1.2 points) and SPR-VLA over MolmoAct (+2.3 points). 
Instead of relying on exact perturbation timing, the online variant infers the timing using proprioceptive history. 
Despite the absence of perturbation-aligned recovery
timing, the online-triggered variant achieves 64.4\%
success, improving the frozen UD-VLA baseline by
8.1 percentage points.
The remaining gap to controlled recovery (74.0\%)
highlights the importance of accurate recovery-entry
timing.


On the standard LIBERO, where no external failures are injected, B2FF consistently outperforms UD-VLA across every reported suite, increasing the average success rate from 91.3\% to 93.7\%. 
This shows that our recovery interface does not degrade the overall performance during nominal execution. Instead, it suggests that familiar-future conditioning effectively mitigates natural deviations that arise during standard rollouts. 
Together, these results indicate that conditioning a frozen foresight-driven VLA on a recoverability-selected familiar future improves robustness to execution failures, bypassing the need to fine-tune the VLA itself.

\begin{table*}[t]
\centering
\footnotesize
\setlength{\tabcolsep}{3.2pt}
\renewcommand{\arraystretch}{1.03}

\begin{tabular}{@{}lccccc@{\hspace{0.8em}}ccccc@{}}
\toprule[1.1pt]
& \multicolumn{5}{c}{Failure-injected LIBERO}
& \multicolumn{5}{c}{Standard LIBERO} \\
\cmidrule(lr){2-6}
\cmidrule(lr){7-11}
Method
& Avg. & Object & Spatial & Goal & Long
& Avg. & Object & Spatial & Goal & Long \\
\midrule
DP$^\dagger$~\cite{diffusionpolicy2023}
& 26.9 & 37.5 & 24.2 & 42.5 & 3.3
& 73.3 & 87.2 & 72.8 & 83.8 & 49.2 \\

DCDP$^\dagger$~\cite{dcdp2026}
& 28.1 & 37.5 & 20.8 & 47.5 & 6.7
& 74.7 & 85.2 & 73.4 & 85.0 & 55.0 \\

MolmoAct~\cite{lee2025molmoact}
& 48.3 & 43.3 & 38.3 & 60.8 & 50.8
& 86.8 & \textbf{95.4} & 87.0 & 87.6 & 77.2 \\

SPR-VLA~\cite{spr2026}
& 50.6 & 44.2 & 50.8 & 56.7 & 50.8
& 91.8 & \textbf{95.4} & 93.2 & \textbf{93.2} & 85.4 \\
\midrule
UD-VLA$^\dagger$~\cite{udvla2026}
& 56.3 & 52.5 & 58.3 & 58.3 & 55.8
& 91.3 & 93.2 & 94.2 & 88.8 & 88.8 \\

\methodshort{}
& \textbf{74.0} & \textbf{69.2} & \textbf{65.8} & \textbf{73.3} & \textbf{87.5}
& \textbf{93.7} & 94.8 & \textbf{95.8} & 91.6 & \textbf{92.6} \\

\methodshort{} (online trigger)
& 64.4 & 61.7 & 60.0 & 72.5 & 63.3
& 92.0 & 93.4 & 95.0 & 89.0 & 90.4 \\
\bottomrule[1.1pt]
\end{tabular}

\caption{
Quantitative results on the failure-injected and standard LIBERO. Values represent final task success rates (\%).
Failure-injected LIBERO evaluates recovery under controlled conditions, whereas Standard LIBERO involves no external failure injection.
For the Standard LIBERO columns, \(^{\dagger}\) denotes values reproduced in our implementation.
Failure-injected evaluations consist of 120 episodes per suite, balanced across three failure types.
In failure-injected evaluation, B2FF uses perturbation-aligned recovery timing to isolate milestone selection; B2FF (online trigger) estimates recovery timing from proprioceptive history.
}
\label{tab:main_summary}
\end{table*}
\begin{table*}[t]
\centering
\scriptsize
\setlength{\tabcolsep}{3.0pt}
\renewcommand{\arraystretch}{1.08}

\begin{minipage}[t]{0.49\textwidth}
\centering
\textbf{(a) Training objective}\\[0.3em]
\begin{tabular}{lccc|rrrr}
\toprule
Variant & Proxy & Sup. & FT & Grip. & Obj. & Lay. & All \\
\midrule
Scratch sup.
& -- & \checkmark & --
& 62.5 & 67.5 & 60.0 & 63.3 \\

Proxy sup.
& \checkmark & \checkmark & --
& 67.5 & 67.5 & 60.0 & 65.0 \\

Weak RL
& \checkmark & \checkmark & weak RL
& 67.5 & 70.0 & 60.0 & 65.8 \\

RL + Pos. BC
& \checkmark & \checkmark & pos. BC
& 67.5 & 70.0 & 60.0 & 65.8 \\

\rowcolor{B2FFBg}
\textbf{B2FF}
& \checkmark & \checkmark & Huber+LE+BC
& \textbf{67.5} & \textbf{72.5} & \textbf{67.5} & \textbf{69.2} \\
\bottomrule
\end{tabular}
\end{minipage}
\hfill
\begin{minipage}[t]{0.49\textwidth}
\centering
\textbf{(b) Selector input}\\[0.3em]
\begin{tabular}{lccc|rrrr}
\toprule
Variant & Obs. & Cand. & Hist. & Grip. & Obj. & Lay. & All \\
\midrule
Obs. only
& \checkmark & -- & 0
& 50.0 & 55.0 & 47.5 & 50.8 \\

+ Cand.
& \checkmark & \checkmark & 0
& 60.0 & 65.0 & 60.0 & 61.7 \\

+ 1 hist.
& \checkmark & \checkmark & 1
& 62.5 & 70.0 & 60.0 & 64.2 \\

+ 3 hist.
& \checkmark & \checkmark & 3
& 70.0 & 70.0 & 60.0 & 66.7 \\

\rowcolor{B2FFBg}
\textbf{B2FF}
& \checkmark & \checkmark & 4
& \textbf{67.5} & \textbf{72.5} & \textbf{67.5} & \textbf{69.2} \\
\bottomrule
\end{tabular}
\end{minipage}

\caption{
Selector ablations on failure-injected LIBERO-Object.
FT denotes the fine-tuning stage.
Weak RL uses a reduced-strength one-step RL-style update.
RL + Pos. BC adds behavior-cloning regularization toward positive or best candidates.
The full B2FF selector combines positive BC with Huber critic regression and a lower-entropy actor update.
}
\label{tab:selector_ablation}
\end{table*}

\begin{figure}[t]
\centering
\includegraphics[width=1.0\linewidth]{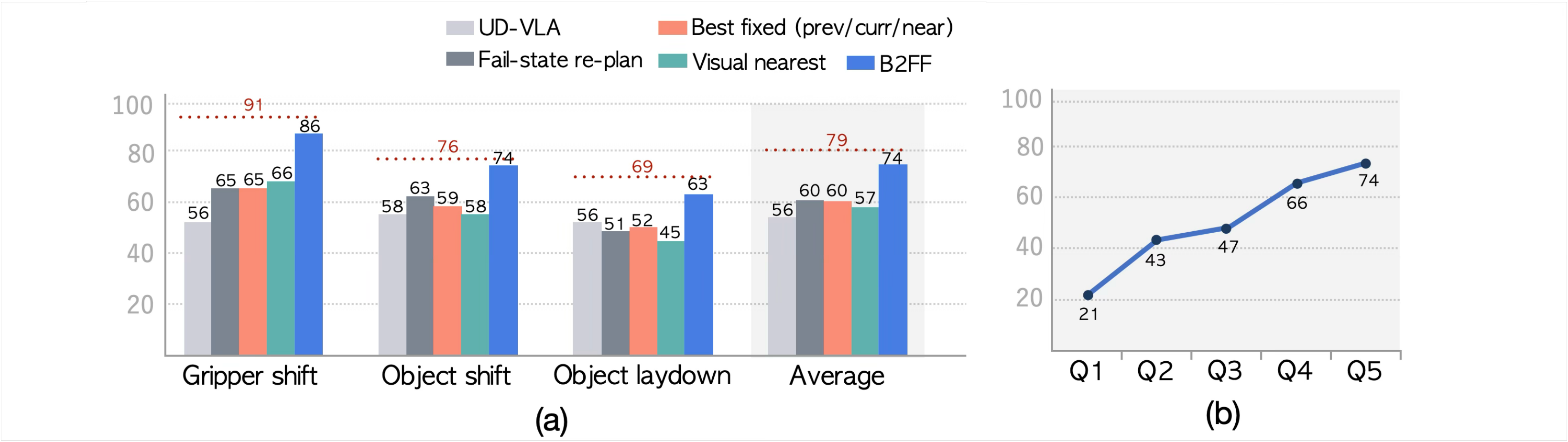}
\caption{
\textbf{Failure-injected LIBERO analysis.}
(a) Selection-rule comparison by injected failure type. Dashed lines show the candidate-selection upper bound within the familiar future bank, and `Best fixed' represents the strongest fixed-anchor baseline among previous, current, and near.
(b) Selector score analysis on failure-injected LIBERO-Object. Candidates are grouped into score quintiles, illustrating that higher selector scores positively correlate with higher recovery success.
}
\label{fig:exp_1}
\end{figure}

\begin{figure}[t]
\centering
\vspace{-1ex}
\includegraphics[width=1.0\linewidth]{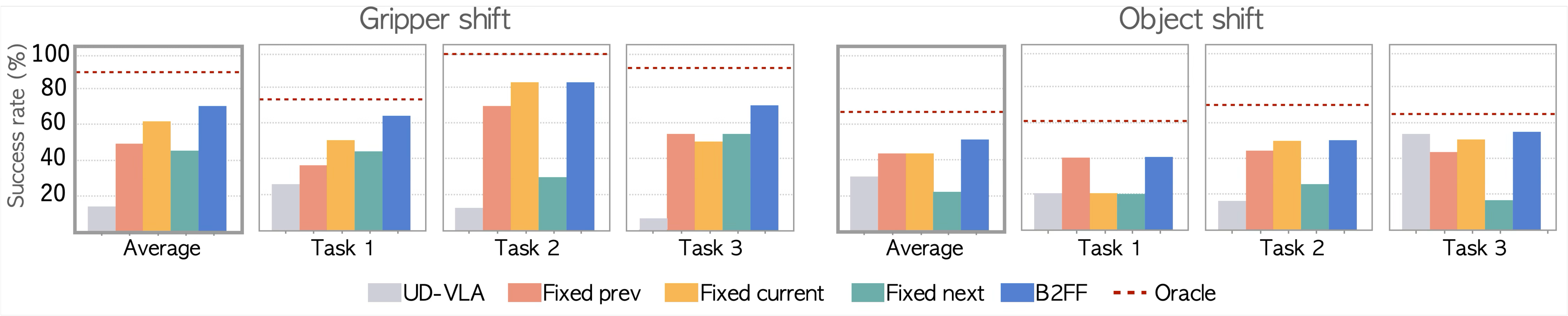}
\vspace{-4ex}
\caption{
Real-world recovery performance by task and failure type.
}
\vspace{-1ex}
\label{fig:exp_2}
\end{figure}

\subsection{In-Depth Analysis}
\label{subsec:in-depth-analysis}

\paragraph{Effectiveness of recoverability-aware milestone selection.}
Figure~\ref{fig:exp_1}a summarizes the selection-rule ablation by injected failure type. 
B2FF achieves the strongest performance across all three failure types, improving average success over failed-state re-planning by +12.3 percentage points. 
This indicates that re-planning from the off-trajectory observation can be unreliable, while a selected milestone provides a more stable action-guiding condition. 
Fixed progress rules and visual-nearest matching are weaker or inconsistent, showing that nominal progress or image similarity alone is not a reliable proxy for recoverability. 
The remaining gap to the candidate oracle suggests that the familiar bank often contains viable anchors, but recovery still depends on selecting the right one.

\begin{figure*}[t]
    \centering

    \begin{minipage}[t]{0.48\textwidth}
        \centering
        \includegraphics[
            width=0.97\linewidth,
            height=3.4cm,
            keepaspectratio
        ]{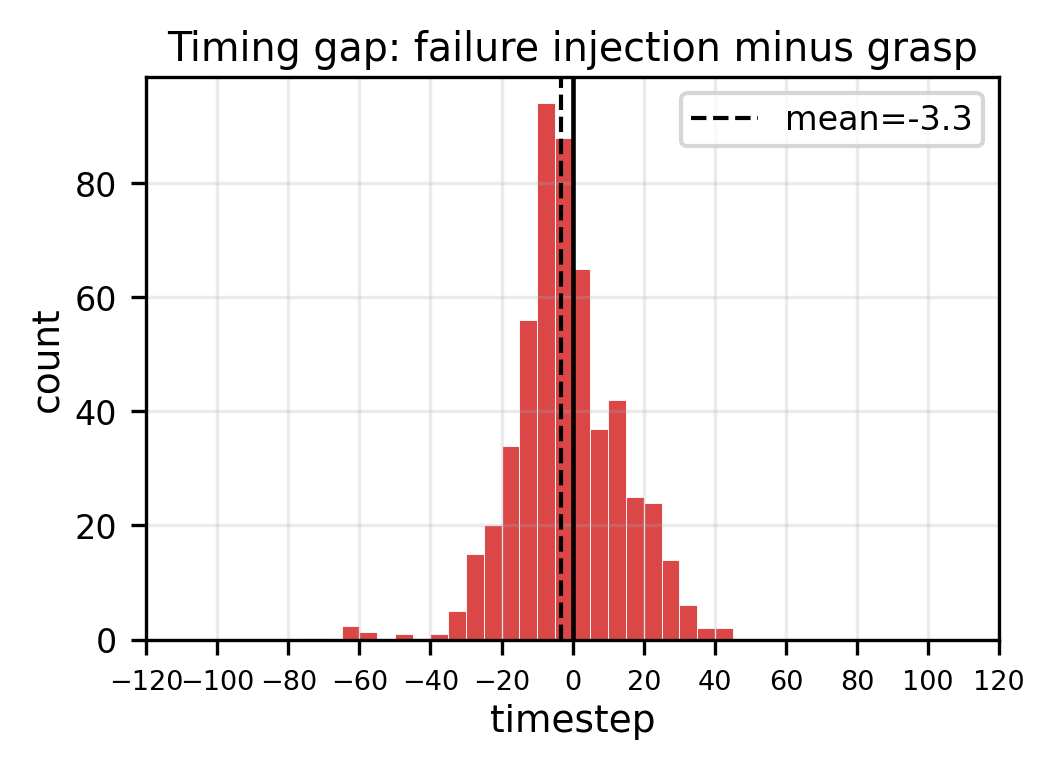}

    \end{minipage}
    \hfill
    \begin{minipage}[t]{0.48\textwidth}
        \centering
        \includegraphics[
            width=0.97\linewidth,
            height=3.4cm,
            keepaspectratio
        ]{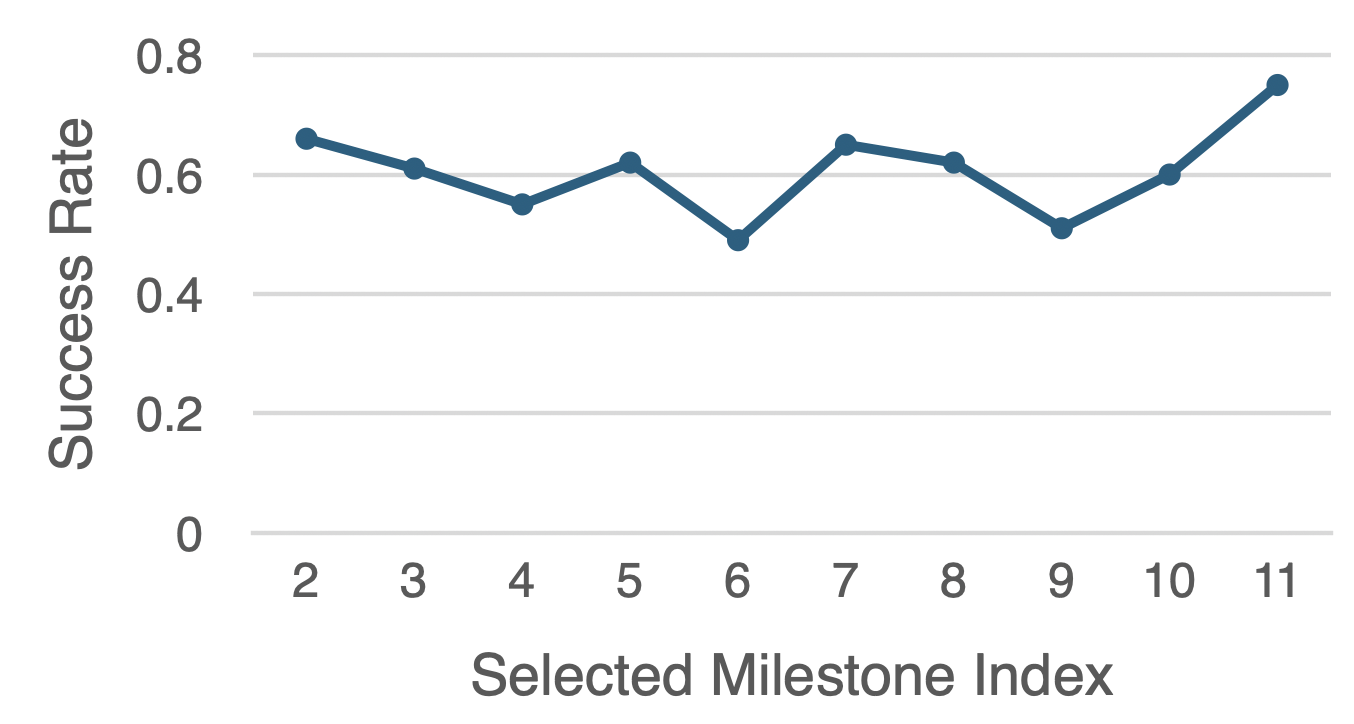}

    \end{minipage}

    \vspace{3pt}
    \caption{
        (a) Distribution of failure-injection timing. 
        (b) Recovery success rate grouped by the
        selected milestone index.
    }
    \label{fig:recovery_diagnostics}
\end{figure*}

We further examine recovery success across selected
milestone indices in Figure~\ref{fig:recovery_diagnostics}(b).
Success varies across the imagined trajectory without
a clear monotonic relationship with milestone index,
suggesting that recovery performance is not simply
determined by selecting a later progress point.
This observation further motivates recoverability-aware
selection over fixed progress-based heuristics.

\begin{table*}[t]
    \centering
    \small
    \renewcommand{\arraystretch}{1.1}

    \begin{minipage}[t]{0.48\textwidth}
        \vspace{0pt}
        \centering
        \textbf{(a) Real-past rewind vs. B2FF}\par
        \vspace{4pt}

        \begin{tabular*}{\linewidth}{
            @{\extracolsep{\fill}}lcc@{}
        }
            \toprule
            Suite & Real-past & B2FF \\
            \midrule
            Object  & 51.7 & \textbf{69.2} \\
            Spatial & 57.5 & \textbf{65.8} \\
            Goal    & 67.5 & \textbf{73.3} \\
            Long    & 72.5 & \textbf{87.5} \\
            \midrule
            \textbf{Average} & 62.3 & \textbf{74.0} \\
            \bottomrule
        \end{tabular*}
    \end{minipage}%
    \hfill
    \begin{minipage}[t]{0.48\textwidth}
        \vspace{0pt}
        \centering
        \textbf{(b) VLA fine-tuning}\par
        \vspace{4pt}

        \begin{tabular*}{\linewidth}{
            @{\extracolsep{\fill}}lcc@{}
        }
            \toprule
            Method & Fail.-inj. & Standard \\
            \midrule
            UD-VLA     & 52.5 & 93.2 \\
            Success-BC & 34.2 & 73.0 \\
            Reward-BC  & 25.0 & 66.0 \\
            DPO-style  & 48.3 & 79.0 \\
            \midrule
            \textbf{B2FF} & \textbf{69.2} & \textbf{94.8} \\
            \bottomrule
        \end{tabular*}
    \end{minipage}

    \vspace{5pt}
    \caption{
        Additional recovery baseline comparisons.
        (a) Real-past rewind versus B2FF on failure-injected LIBERO.
        (b) Direct VLA fine-tuning on LIBERO-Object under
        the same recovery-rollout budget.
        Values denote task success rates (\%).
    }
    \label{tab:additional_baselines}
    \vspace{-4pt}
\end{table*}

\paragraph{Imagined futures versus real past observations.}
To investigate whether recovery benefits simply from conditioning on familiar observations, we additionally evaluate a real-past rewind baseline that uses previously observed states as recovery anchors.
Table~\ref{tab:additional_baselines}(a) compares B2FF against this baseline across four LIBERO suites.
While past observations provide familiar visual conditions, they are limited to states already encountered during execution.
In contrast, the pre-imagined future bank provides diverse candidate recovery targets, enabling rollback, retry, and skip-ahead hypotheses.
These results suggest that access to familiar future conditions offers advantages beyond reusing previously visited states.

\paragraph{Recovery through visual conditioning versus direct VLA fine-tuning.}
To examine whether recovery supervision is better utilized by adapting the action generator, we compare B2FF against direct VLA fine-tuning baselines under the same recovery-rollout budget on LIBERO-Object.
Table~\ref{tab:additional_baselines}(b) summarizes the comparison against Success-BC, Reward-BC, and DPO-style adaptation.
The evaluated fine-tuning approaches reduce performance on Standard LIBERO, whereas B2FF preserves strong nominal-task performance.
These results suggest that, under limited recovery supervision, selecting an appropriate visual condition can be more effective than directly adapting the low-level action generator.

\paragraph{Effectiveness of the learned selector.}
Table~\ref{tab:selector_ablation} analyzes the selector on failure-injected LIBERO-Object. 
The training ablation shows that selector performance improves when proxy temporal initialization is followed by the full fine-tuning variant, increasing overall success rate. 
The intermediate variants provide only partial gains, suggesting that positive-BC regularization alone is insufficient without the more stable Huber critic regression and lower-entropy actor update. 
The input ablation shows that candidate-relative context is essential: observation-only scoring is substantially weaker, while adding candidate milestones and recent history enables the selector to judge which familiar future is recoverable from the current failure context. 
Figure~\ref{fig:exp_1}b further validates that the learned scores are meaningful: when candidates are binned by selector-score quintiles on failure-injected LIBERO-Object, recovery success increases from low-score to high-score bins. 

\begin{figure}[t]
\centering
\includegraphics[width=1.0\linewidth]{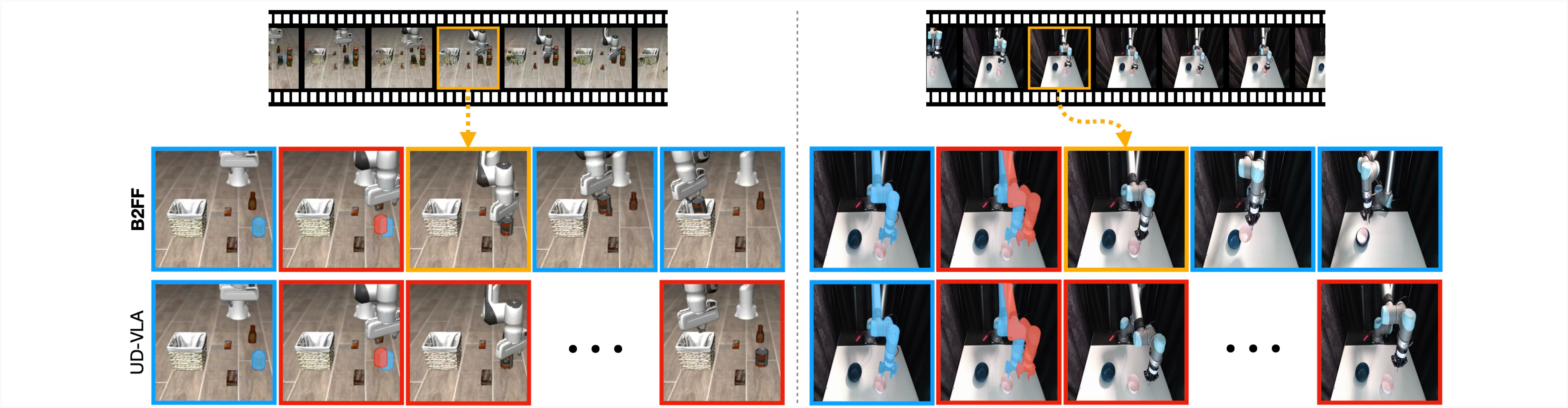}
\vspace{-3ex}
\caption{
Qualitative recovery examples.
Left: simulation object-shift perturbation.
Right: real-world gripper-shift perturbation.
B2FF selects a pre-imagined milestone from the top familiar future bank and uses it as the fixed future-image subgoal, while UD-VLA re-predicts from the off-trajectory observation.
}
\label{fig:qual}
\end{figure}

\paragraph{Qualitative analysis.}
Figure~\ref{fig:qual} shows that, after simulation object-shift and real-world gripper-shift perturbations, B2FF selects a pre-imagined milestone that acts as a familiar future anchor, whereas UD-VLA re-predicts from the off-trajectory observation and can follow an unreliable recovery path.

\subsection{Real-World Experiment}
\label{subsec:real-world}
We further evaluate B2FF on a real robot setup across three tasks: object stacking, pick-and-place, and drawer closing with object placement. 
The underlying real-world VLA policy is trained on 600 demonstration trajectories. 
For the recovery module, we initialize the milestone selector from LIBERO training and fine-tune only the selector on 35 real-world recovery groups. 
We then evaluate on 90 separate real-world recovery trials with two injected failures: gripper shift and target-object shift. 
Figure~\ref{fig:exp_2} shows that B2FF achieves the strongest real-world recovery performance. 
Across the 90 evaluation trials, B2FF reaches 61.1\% overall success, outperforming both the base UD-VLA policy and fixed-anchor baselines, with consistent gains under both failures. 
These results suggest that the familiar-future recovery mechanism observed in simulation also transfers to real robot settings with lightweight selector tuning.
Notably, the LIBERO-trained milestone selector achieves 55.6\% success without real-world recovery tuning. Fine-tuning the selector using only 35 real-world recovery contexts improves success to 61.1\%. This suggests that the learned recoverability ranking transfers to physical manipulation settings, while lightweight domain-specific adaptation provides additional benefits.

\section{Limitations}

B2FF targets recoverable off-trajectory deviations in which the task remains physically feasible: relevant objects remain observable and reachable, and the required manipulation skill is already available in the underlying VLA. 
This includes natural rollout errors, contact-induced object slips, mild misalignments, and externally injected perturbations. 
B2FF does not address semantic or irreversible failures, such as incorrect task understanding, wrong object grounding, missing objects, severe perception failures, workspace exits, irreversible environment changes, or skills absent from the VLA. B2FF also assumes a foresight-driven VLA with a settable future-image subgoal interface. 
The familiar future bank need not pixel-match the failed observation, but it must contain a milestone that provides a useful action-guiding condition for the current failed state; otherwise, recovery may fail. 
Our implementation constructs a single pre-execution bank, selects one milestone for a fixed recovery window, and depends on recovery-trigger timing. 
Adaptive bank construction, closed-loop re-selection, repeated-trigger handling, and stronger online trigger estimation remain future work.

\section{Conclusion}

We presented \textsc{Back to the Familiar Future} (B2FF), a recovery framework for frozen foresight-driven VLA policies. 
B2FF reframes recovery as selecting a pre-imagined future-image subgoal rather than re-planning directly from an unfamiliar off-trajectory observation or modifying the low-level action generator. 
Across failure-injected and standard LIBERO, and in real-world recovery trials, B2FF shows that familiar visual conditions can guide frozen VLA policies back toward successful execution. 
These results suggest that visual-condition selection is a practical mechanism for recovery in foresight-driven VLAs.
\clearpage

\acknowledgments{This work was partly supported by grants from the IITP (RS-2021-II211343-GSAI/10\%, RS-2022-II220951-LBA/10\%, RS-2022-II220953-PICA/10\%, RS-2026-25553157-MIACC/10\%), the NRF (RS-2024-00353991-SPARC/10\%, RS-2023-00274280-HEI/10\%), the KEIT (RS-2025-25453780/10\%), the KIAT (RS-2025-25460896/10\%), and the IITP (RS-2026-25507282/20\%), founded by the Korean government.  }


\bibliography{example}  

\appendix
\clearpage
\section*{Appendix}

\appendix
\setcounter{section}{0}
\setcounter{subsection}{0}

\renewcommand{\thesection}{\Alph{section}}
\renewcommand{\thesubsection}{\thesection.\arabic{subsection}}
\renewcommand{\thesubsubsection}{\thesubsection.\arabic{subsubsection}}

\section{B2FF Implementation Details}
\label{app:b2ff_implementation}

\noindent\textbf{Frozen inference interface.}
As described in the main text, B2FF uses the frozen foresight-driven VLA in two modes:
joint subgoal--action generation during nominal execution,
$\pi_\theta(v_t,a_t \mid I,o_t)$, and action-only denoising during recovery,
$\pi_\theta(a_t \mid I,o_t;\; v_t \leftarrow v^\star)$. Throughout all appendix
experiments and algorithms, all calls to $\pi_\theta$ use frozen VLA weights: B2FF does not
update the visual tokenizer, future-image generator, action denoiser, or any other VLA backbone
component.


\begin{algorithm}[H]
\caption{\textsc{B2FF} Inference}
\label{alg:b2ff_inference}
\begin{algorithmic}[1]
\Require Frozen VLA $\pi_\theta$, selector $F_\phi$, instruction $I$, initial observation $o_0$,
bank size $M_0$, offsets $\Delta$, recovery window $W$, decision horizon $T$
\Ensure Executed trajectory

\State $B \gets \textsc{BuildBank}(\pi_\theta,I,o_0,M_0)$
\State $r \gets 0$, \quad $v^\star \gets \emptyset$

\For{decision chunk $t=1,\ldots,T$}
    \State Observe $o_t,p_t$ and update history $H_t$ and $p_{1:t}$

    \If{$r=0$}
        \State $f \gets \textsc{TriggerIndex}(t,p_{1:t})$
        \If{$f \neq \emptyset$}
            \State $(B,C_f) \gets \textsc{BuildCandidates}(B,f,\Delta)$
            \State $v^\star \gets
            \arg\max_{\tilde v\in C_f}
            F_\phi(\tilde v \mid o_t,H_t,C_f)$
            \State $r \gets W$
        \EndIf
    \EndIf

    \If{$r>0$}
        \State $a_t \sim \pi_\theta(a \mid I,o_t;\; v_t \leftarrow v^\star)$
        \State $r \gets r-1$
    \Else
        \State $(v_t,a_t) \sim \pi_\theta(v,a \mid I,o_t)$
    \EndIf

    \State Execute action chunk $a_t$

\EndFor
\end{algorithmic}
\end{algorithm}

\subsection{Familiar Future Bank and Candidate Construction}
\label{app:familiar_future_bank}

Before executing any robot action, B2FF constructs an initial familiar future bank from the clean
initial observation $o_0$. We set $\tilde v_0=o_0$ and recursively query the future-image marginal
of the frozen VLA:
\[
    \tilde v_m \sim \pi_\theta(v \mid I,\tilde v_{m-1}),
    \qquad m=1,\ldots,M_0 .
\]
The initial bank is $B=\{\tilde v_1,\ldots,\tilde v_{M_0}\}$. No actions are executed during this imagination rollout, so the bank contains familiar
future imagined from the clean initial state rather than futures re-predicted from a
failed observation. We denote this pre-execution imagination procedure by
\textsc{BuildBank} in Algorithm~\ref{alg:b2ff_inference}.

When recovery begins at chunk-level index $f$, B2FF forms a local candidate set using offset set
$\Delta$. This procedure is denoted by
\textsc{BuildCandidates}$(B,f,\Delta)$ in Algorithm~\ref{alg:b2ff_inference}. Before returning the
candidate set, we check the largest requested candidate index
\[
    m_{\max}=\max_{\delta\in\Delta}(f+\delta).
\]
If $m_{\max}$ exceeds the current bank length, we deterministically extend the bank by continuing
imagination from the last available bank image until the required index exists. This tail extension
is used only to complete the local candidate set near the end of the initial bank. It does not query
the VLA from the failed observation and does not execute actions. Lower out-of-range indices are
discarded rather than extended backward.

After this optional tail extension, the local candidate set is
\[
    C_f =
    \{\tilde v_{f+\delta}
    \mid
    \delta\in\Delta,\;
    1\leq f+\delta\leq |B|
    \}.
\]

\subsection{B2FF Inference Algorithm}
\label{app:b2ff_algorithm}

\noindent
\textsc{BuildBank} denotes the pre-execution imagination rollout in
Sec.~\ref{app:familiar_future_bank}. \textsc{BuildCandidates} returns the local candidate set and
the possibly extended bank; when an upper-end candidate index exceeds the current bank length,
it deterministically extends the bank from the last available familiar milestone, as described in
Sec.~\ref{app:familiar_future_bank}. \textsc{TriggerIndex} returns either a chunk-level recovery
index $f$ or $\emptyset$. In controlled failure-injected evaluation, it maps the injected perturbation
time to the corresponding decision-chunk index. In the online-triggered variant, it maps the first
detector threshold crossing to the corresponding decision-chunk index using the same action-chunk
convention. If no new trigger is active, \textsc{TriggerIndex} returns $\emptyset$ and B2FF remains
in nominal execution.
The counter $r$ stores the number of remaining recovery chunks. Thus, $r>0$ is equivalent to
recovery mode, and $r=0$ is equivalent to nominal mode. Once $v^\star$ is selected, B2FF keeps it
fixed as the future-image subgoal for $W$ consecutive action-generation calls. 
\section{Recovery-Mode Entry and Online Trigger}
\label{app:recovery_entry_online_trigger}

\subsection{Controlled Recovery Timing}
\label{app:controlled_recovery_timing}

We use a controlled recovery-entry protocol to evaluate the recovery mechanism independently of
online failure detection. The recovery-entry signal specifies only when B2FF switches from nominal
execution to recovery mode. It does not reveal which familiar milestone should be selected, does not
provide candidate-wise recovery outcomes, and does not modify the recovery policy.

In failure-injected LIBERO, the recovery-entry signal is given by the known injected perturbation
time. B2FF enters recovery mode at the corresponding decision-chunk index; if the perturbation
occurs between two decision chunks, recovery begins at the next chunk-level decision.

In Standard LIBERO, where no external perturbation is injected, we use the same controlled-entry
protocol with a human-provided annotation of the first clear off-trajectory or failure-onset point.
If no such deviation is annotated, the policy remains in nominal execution. This annotation serves
the same role as the perturbation time in the failure-injected setting: it determines only the recovery
entry point, while milestone selection and recovery action generation are still performed by each
method according to its own rule.

This protocol isolates the quality of the recovery-time visual condition from the separate problem of
detecting failures online. The online-triggered variant removes this controlled entry signal and
instead estimates the recovery-entry time from proprioceptive history.

\subsection{Online-Triggered Variant}
\label{app:online_detector}

The online-triggered variant estimates the recovery-entry time from history available during
execution. When a trigger fires, \textsc{TriggerIndex} maps the first threshold crossing to the
corresponding decision-chunk recovery index. If no trigger fires, B2FF remains in nominal execution
mode.

Table~\ref{tab:online_trigger_metrics} summarizes the trigger metric pool used in the online-triggered
variant. The trigger determines only the recovery-entry time; milestone selection follows the same
candidate-selection interface as in the controlled setting.
\begin{table}[t]
\centering
\caption{
Online-trigger metric pool. The trigger uses proprioceptive history and task-progress signals; no
camera observations, familiar future candidates, selector scores, or candidate-wise recovery rollouts
are used by the trigger.
}
\label{tab:online_trigger_metrics}
\footnotesize
\setlength{\tabcolsep}{4pt}
\renewcommand{\arraystretch}{1.08}
\begin{tabularx}{\linewidth}{@{}p{0.19\linewidth}p{0.28\linewidth}X@{}}
\toprule
Metric & Signal type & Trigger evidence \\
\midrule
\textsc{TJ}
& Trajectory deviation
& Large next-chunk proprioceptive prediction error. \\

\textsc{GP}
& Gripper state
& Gripper width or command pattern outside the calibrated successful range. \\

\textsc{DL}
& End-effector stall
& Sustained low Cartesian motion despite continued control. \\

\textsc{TaskTop4}
& Task-progress consistency
& Intended task progress becomes inconsistent with the proprioceptive stream. \\

\textsc{Consistency}
& Motion consistency
& Repeated sign reversals or unstable short-horizon motion patterns. \\

\textsc{KNN\mbox{-}State}
& State novelty
& Large nearest-neighbor distance from successful proprioceptive states. \\
\bottomrule
\end{tabularx}
\end{table}

\subsection{Calibration and Runtime Protocol}
\label{app:online_trigger_calibration}

Trigger thresholds are calibrated using successful demonstrations or successful replay trajectories
and are frozen before evaluation. The calibration stage may fit lightweight proprioceptive models or
reference banks for the signal groups in Table~\ref{tab:online_trigger_metrics}, but it does not use
candidate-wise recovery labels or held-out failure-injected test episodes.

At runtime, the enabled trigger signals are combined by an OR rule, and the earliest threshold
crossing becomes the online recovery-entry event. No trigger thresholds, fitted models, or reference
banks are updated online.

We use one trigger framework with a shared metric pool and split-specific enabled metric sets,
rather than imposing a single global trigger rule on all suites. This is because different suites and
failure modes produce different proprioceptive signatures. The metric pool and calibration protocol
are shared across splits; only the subset of metrics connected to the OR rule changes, and these
choices are fixed before evaluation.

For object failure-injected rollouts, we enable
\[
    \textsc{TJ}
    \;\lor\;
    \textsc{GP}
    \;\lor\;
    \textsc{DL}.
\]
This combination targets trajectory deviation, gripper-state anomalies, and end-effector stalls. For
most other splits, we enable
\[
    \textsc{TaskTop4}
    \;\lor\;
    \textsc{GP}
    \;\lor\;
    \textsc{Consistency}
    \;\lor\;
    \textsc{KNN\mbox{-}State}.
\]
This combination targets task-progress inconsistency, gripper-state anomalies, unstable short-horizon
motion, and proprioceptive state novelty. Depending on the split, thresholds are calibrated from
either expert demonstrations or medium-quality successful replays.

Our reported failure-injected benchmark contains one external perturbation per episode. Accordingly,
the online-triggered result evaluates the first recovery-entry decision after a detected deviation rather
than a full repeated-trigger policy. During an active recovery window, B2FF keeps the selected
milestone fixed and does not switch targets mid-window.
\section{Recoverability-Aware Selector Architecture}
\label{app:selector_architecture}

\subsection{Selector Interface and Proxy-VQ Features}
\label{app:selector_inputs}
\label{app:frozen_visual_tokenizer}

At recovery entry index $f$, the selector scores candidate familiar milestones under the recovery
context $(o_f,H_f,C_f)$, where $o_f$ is the current failure observation, $H_f$ is a short observation
history, and $C_f$ is the local candidate set constructed in
Sec.~\ref{app:familiar_future_bank}. The final model uses up to four history frames and up to five
candidate milestones from the default offset set $\Delta=\{-1,0,+1,+2,+4\}$.

For each valid candidate $c_i\in C_f$, the selector outputs a scalar recoverability logit
\[
    s_i = F_\phi(c_i \mid o_f,H_f,C_f).
\]
The score is used only for ranking candidate milestones, not as a calibrated success probability.
At test time, B2FF selects the highest-scoring valid candidate and fixes it as the future-image
subgoal for recovery.

The selector uses pre-tokenized visual tokens from the frozen Emu3/UD-VLA visual tokenizer. Each
frame contains two camera views: a full-view image with $25\times25=625$ visual tokens and a
wrist-view image with $10\times10=100$ visual tokens. Each valid visual token id is mapped through
the frozen Emu3 VQ codebook to a 4-dimensional embedding. We then flatten and concatenate the
two-view codebook embeddings to obtain a frame-level proxy-VQ feature:
\[
    z(x)\in\mathbb{R}^{(625+100)\cdot 4}
    =
    \mathbb{R}^{2900}.
\]
Thus, every observation, history frame, and candidate milestone is represented as a
2900-dimensional continuous proxy-VQ feature.


\subsection{Projectors and Candidate-Context Encoder}
\label{app:step_pair_projectors}
\label{app:perceiver_attention}

Each proxy-VQ frame feature is first mapped to a 256-dimensional step representation:
\[
    h_x = g_{\mathrm{step}}(z(x))\in\mathbb{R}^{256}.
\]
The step projector is implemented as
\[
    \mathrm{LayerNorm}(2900)
    \rightarrow
    \mathrm{Linear}(2900,512)
    \rightarrow
    \mathrm{GELU}
    \rightarrow
    \mathrm{Linear}(512,256)
    \rightarrow
    \mathrm{GELU}.
\]

To compare the current failure observation against a history frame or candidate milestone, we use a
pair projector. For any frame $x\in H_f\cup C_f$, the pair feature is
\[
    p_x = g_{\mathrm{pair}}([h_{o_f};h_x])\in\mathbb{R}^{256}.
\]
The final implementation uses only the concatenation $[h_{o_f};h_x]$. The pair projector is
implemented as
\[
    \mathrm{LayerNorm}(512)
    \rightarrow
    \mathrm{Linear}(512,512)
    \rightarrow
    \mathrm{GELU}
    \rightarrow
    \mathrm{Linear}(512,256)
    \rightarrow
    \mathrm{GELU}.
\]

During the final RL-style selector fine-tuning stage, the step projector is frozen. The pair projector
is updated with a reduced learning rate, set to $0.3\times$ the main selector learning rate.

The selector then uses a Perceiver-style candidate-context encoder. All selector hidden states have
dimension 256. The context encoder uses 16 learnable latent slots, 8 attention heads, dropout 0.1,
and an MLP ratio of 4.0, corresponding to a 1024-dimensional feed-forward hidden layer.

Let
\[
    P_H=\{p_x \mid x\in H_f\}
\]
denote the history pair tokens. The learnable context latents
$\Lambda_0\in\mathbb{R}^{16\times256}$ first cross-attend to the history pair tokens and are then
processed by two latent self-attention blocks:
\[
    \Lambda =
    \mathrm{SelfAttnBlock}_{2}
    \left(
        \mathrm{SelfAttnBlock}_{1}
        \left(
            \mathrm{CrossAttn}(\Lambda_0,P_H)
        \right)
    \right).
\]

Candidate pair tokens are processed by a candidate encoder. Let
\[
    Q_C=\{p_{c_i}\mid c_i\in C_f\}
\]
denote the candidate pair tokens. The candidate encoder applies two blocks, each consisting of
candidate self-attention followed by cross-attention from candidates to the context latents:
\[
    Q_C^{\mathrm{out}}
    =
    \mathrm{CrossAttnBlock}_{2}
    \left(
        \mathrm{SelfAttnBlock}_{2}
        \left(
            \mathrm{CrossAttnBlock}_{1}
            \left(
                \mathrm{SelfAttnBlock}_{1}(Q_C),
                \Lambda
            \right)
        \right),
        \Lambda
    \right).
\]
This architecture allows candidate milestones to be compared against one another while also
attending to the recent recovery context.

Invalid or padded candidate slots are masked during candidate attention. Before the final selection,
their logits are set to $-10^9$ so that they cannot be selected.

\subsection{MLP Scoring Head}
\label{app:scoring_head}

For each valid candidate, the candidate-context encoder returns a final representation
$r_i\in\mathbb{R}^{256}$. The scoring head applies layer normalization followed by a scalar linear
head:
\[
    s_i = W_{\mathrm{score}}\,\mathrm{LN}(r_i) + b_{\mathrm{score}}.
\]
The resulting scalar $s_i$ is the recoverability logit used to rank candidate milestones. During
inference, B2FF selects the valid candidate with the largest score and uses it as the fixed
future-image subgoal for the recovery window.

\subsection{Architecture and Candidate-Count Ablations}
\label{app:selector_arch_ablation}

\paragraph{Architecture ablation.}
Table~\ref{tab:perceiver_ablation} compares the final Perceiver-style selector against simpler
candidate scoring architectures. The MLP-only variant scores each candidate independently. The
Concat MLP variant concatenates candidate and context features before scoring. The shared-context
variant builds a single context feature and scores each candidate against it. The full selector uses
Perceiver-style candidate-context attention and achieves the strongest overall recovery performance.

\begin{table}[H]
\centering
\caption{
Selector architecture ablation on failure-injected LIBERO-Object. All variants are evaluated under
the same 120-episode Object recovery protocol. Values are overall task success rates (\%), rounded
to one decimal.
}
\label{tab:perceiver_ablation}
\footnotesize
\setlength{\tabcolsep}{5pt}
\renewcommand{\arraystretch}{0.98}
\begin{tabular}{@{}lc@{}}
\toprule
Variant & Success (\%) \\
\midrule
MLP only & 61.7 \\
Concat MLP & 62.5 \\
Shared context + candidate & 66.7 \\
\textbf{Full Perceiver selector} & \textbf{69.2} \\
\bottomrule
\end{tabular}
\end{table}

\paragraph{Candidate-count ablation.}
Table~\ref{tab:candidate_set_ablation} ablates which familiar milestones are exposed to the
selector. Candidate slots are defined relative to the recovery entry index $f$: previous
($f-1$), current ($f$), near ($f+1$), mid ($f+2$), and far ($f+4$). The final setting uses all five
candidate slots, corresponding to $\Delta=\{-1,0,+1,+2,+4\}$.

\begin{table}[H]
\centering
\caption{
Candidate-count ablation on failure-injected LIBERO-Object. Check marks indicate which familiar
milestone slots are included in the candidate set. Values are overall task success rates (\%), rounded
to one decimal.
}
\label{tab:candidate_set_ablation}
\footnotesize
\setlength{\tabcolsep}{4.5pt}
\renewcommand{\arraystretch}{1.00}
\begin{tabular}{@{}c ccccc c@{}}
\toprule
\#Cand. & Prev. & Curr. & Near & Mid & Far & Success (\%) \\
\midrule
2
& \checkmark & \checkmark & -- & -- & --
& 58.3 \\

3
& \checkmark & \checkmark & \checkmark & -- & --
& 59.2 \\

4
& \checkmark & \checkmark & \checkmark & \checkmark & --
& 60.0 \\
&
\checkmark & \checkmark & \checkmark & -- & \checkmark
& 66.7 \\
&
\checkmark & \checkmark & -- & \checkmark & \checkmark
& 67.5 \\
&
\checkmark & -- & \checkmark & \checkmark & \checkmark
& 64.2 \\
&
-- & \checkmark & \checkmark & \checkmark & \checkmark
& 65.8 \\

\textbf{5}
& \checkmark & \checkmark & \checkmark & \checkmark & \checkmark
& \textbf{69.2} \\
\bottomrule
\end{tabular}
\end{table}
\section{Selector Training}
\label{app:selector_training}

\subsection{Training Pipeline Overview}
\label{app:selector_training_overview}

The recoverability-aware selector is trained offline while the foresight-driven VLA remains frozen.
The training pipeline has three stages. First, proxy temporal initialization learns progress-aware step
and pair projectors from successful trajectories. Second, counterfactual recovery rollouts produce
candidate-wise success labels for injected failure contexts. Third, the selector is trained with a
supervised warm-start and then refined with a one-step actor-critic-style offline objective.

\subsection{Proxy Temporal Initialization}
\label{app:proxy_temporal_initialization}

Training the selector directly from a small number of counterfactual recovery groups is
sample-inefficient. We therefore initialize the step and pair projectors with a proxy temporal task
before supervised recoverability training. The proxy task asks whether an observation and a
candidate future milestone are temporally close or far apart, producing an embedding space that is
aware of nominal task progress.

\paragraph{Data.}
We use successful trajectories from the existing LIBERO training split for
LIBERO-Long, LIBERO-Goal, LIBERO-Object, and LIBERO-Spatial, rather than collecting an
additional dataset for this proxy task. From these trajectories, we extract observation--future pairs
indexed by trajectory steps $a$ and $b$. Since B2FF makes recovery decisions at action-chunk
resolution, we convert the temporal gap to a decision-chunk gap,
\[
    g_{\mathrm{chunk}}
    =
    \left\lceil
    \frac{|b-a|}{L}
    \right\rceil,
\]
where $L=10$ is the action-chunk length used in our experiments. We stratify samples into five
coarse chunk-gap buckets,
\[
    [1,2],\quad [3,4],\quad [5,8],\quad [9,16],\quad >16 .
\]
These approximately logarithmic buckets cover local, medium-range, and long-range temporal
relationships while reducing the dominance of frequent short-gap pairs. This produces 190{,}156
samples, split into 80\% training, 10\% validation, and 10\% test sets.

\paragraph{Model.}
The proxy model uses the same frozen proxy-VQ frame features $z(x)$ defined in
Sec.~\ref{app:frozen_visual_tokenizer}. A shared step projector maps each frame feature to a
256-dimensional representation,
\[
    s_x = g_{\mathrm{step}}(z(x)) \in \mathbb{R}^{256}.
\]
Given an observation at step $a$ and a candidate future milestone at step $b$, a pair projector maps
the concatenated step features to a pair representation:
\[
    z_{ab} = g_{\mathrm{pair}}([s_a;s_b]).
\]
An auxiliary gap head predicts a 5-way temporal-gap bucket from $z_{ab}$. After proxy
pre-training, the gap head is discarded and the step and pair projectors initialize the downstream
selector.

\paragraph{Loss.}
The proxy objective combines a time-contrastive triplet loss and a temporal-gap classification loss:
\[
    L_{\mathrm{proxy}}
    =
    L_{\mathrm{TCN}}
    +
    \beta_{\mathrm{gap}} L_{\mathrm{gap}},
    \qquad
    \beta_{\mathrm{gap}}=0.3.
\]
For each anchor representation $s_a$, we sample a positive $s_p$ within one decision chunk and a
negative $s_n$ more than ten decision chunks away from the same trajectory. Using cosine distance
\[
    d(u,v)
    =
    1 -
    \frac{u^\top v}{\|u\|_2\|v\|_2},
\]
the time-contrastive loss is
\[
    L_{\mathrm{TCN}}
    =
    \max
    \left(
        0,\,
        d(s_a,s_p)
        -
        d(s_a,s_n)
        +
        \alpha
    \right),
    \qquad
    \alpha=0.2.
\]
The temporal-gap loss is standard cross-entropy over the five chunk-gap buckets.

\paragraph{Optimization and result.}
We train the proxy model for 50 epochs using AdamW with learning rate $1\times10^{-3}$ and a
cosine annealing scheduler. The visual tokenizer and VQ codebook remain frozen. On the held-out
test set, the best checkpoint achieves 94.79\% temporal-gap classification accuracy and 99.16\%
TCN ranking accuracy, where ranking accuracy measures how often
\[
    d(s_a,s_p) < d(s_a,s_n).
\]

\subsection{Counterfactual Rollout Label Collection}
\label{app:counterfactual_rollout_label_collection}

After proxy temporal initialization, we collect offline counterfactual recovery labels. Each training
example is a recovery group defined by one injected failure context,
\[
    x_i = (o_f^i, H_f^i, C_f^i),
\]
where $o_f^i$ is the failure observation, $H_f^i$ is the recent observation history, and $C_f^i$ is
the local candidate set from the familiar future bank.

For each candidate $c\in C_f^i$, we clamp the future-image subgoal to that candidate and roll out
the frozen VLA in action-only denoising mode:
\[
    a_t \sim \pi_\theta(a \mid I,o_t;\; v_t \leftarrow c).
\]
The resulting candidate label is
\[
    y_{i,c}
    =
    \mathbf{1}\{\text{the recovery rollout succeeds}\}.
\]
Thus, each recovery group contains candidate-wise labels
\[
    \mathcal{G}_i
    =
    \{(c,y_{i,c}) \mid c\in C_f^i\}.
\]

We collect 400 injected training failure contexts. With the default offset set
$\Delta=\{-1,0,+1,+2,+4\}$, each group contains at most five valid candidates, giving
\[
    N_{\mathrm{cf}}
    =
    \sum_{i=1}^{400}|C_f^i|
    \leq
    400\times5
    =
    2000
\]
candidate-wise counterfactual rollouts. The exact count can be slightly smaller when lower-index
candidates are invalid near the beginning of a bank. Each valid candidate is evaluated once using a
five-chunk recovery rollout.

\subsection{Supervised Warm-Start Objective}
\label{app:supervised_warm_start}

We first train the selector with supervised candidate-wise objectives. For a recovery group
\[
    \mathcal{G}_f=\{(c,y_c)\mid c\in C_f\},
\]
let
\[
    s_c = F_\phi(c\mid o_f,H_f,C_f)
\]
be the selector score for candidate $c$.

The candidate-wise binary cross-entropy loss is
\[
    L_{\mathrm{BCE}}
    =
    -\frac{1}{|C_f|}
    \sum_{c\in C_f}
    \left[
        y_c \log \sigma(s_c)
        +
        (1-y_c)\log(1-\sigma(s_c))
    \right].
\]
We also use a grouped ranking objective within each recovery group. Let
\[
    Y_f=\sum_{c\in C_f}y_c
\]
be the number of successful candidates. When $Y_f>0$,
\[
    L_{\mathrm{group}}
    =
    -
    \frac{1}{Y_f}
    \sum_{c\in C_f} y_c s_c
    +
    \log
    \sum_{c\in C_f}\exp(s_c).
\]
This objective raises the normalized scores of all successful candidates without forcing an arbitrary
single positive when multiple milestones can recover the task. For all-fail groups, we omit
$L_{\mathrm{group}}$ and train only with the BCE term.

The supervised warm-start objective is
\[
    L_{\mathrm{sup}}
    =
    \lambda_{\mathrm{BCE}} L_{\mathrm{BCE}}
    +
    \lambda_{\mathrm{group}} L_{\mathrm{group}},
    \qquad
    \lambda_{\mathrm{BCE}}=\lambda_{\mathrm{group}}=1.0.
\]

We train the supervised warm-start for 20 epochs using AdamW with learning rate
$3\times10^{-4}$, weight decay $10^{-4}$, batch size 16, and gradient clipping at 1.0. The
checkpoint is selected by validation top-1 recovery success on held-out recovery groups.

\subsection{One-Step Actor-Critic-Style Fine-Tuning}
\label{app:one_step_actor_critic}

We further refine the selector with an offline one-step actor-critic-style objective. Each recovery
context $(o_f,H_f,C_f)$ defines a small discrete action space, where each candidate $c\in C_f$ is
one possible recovery choice. The counterfactual rollout label provides the one-step utility target:
\[
    r_c = y_c.
\]

\paragraph{Critic update.}
We attach two scalar critic heads to the selector candidate representation. For each candidate,
\[
    Q_k(c \mid o_f,H_f,C_f),
    \qquad
    k\in\{1,2\},
\]
predicts the recorded recovery success label. The critic loss is
\[
    L_Q
    =
    \mathbb{E}
    \left[
        \frac{w_f}{|C_f|}
        \sum_{c\in C_f}
        \sum_{k=1}^{2}
        \mathrm{Huber}_{1.0}
        \left(
            Q_k(c\mid o_f,H_f,C_f)-y_c
        \right)
    \right].
\]
\paragraph{Group weighting.}
We use recovery-group weights to emphasize contexts where the candidate choice matters:
\[
w_f =
\begin{cases}
0.5, & \text{all candidates fail},\\
0.5, & \text{all candidates succeed},\\
1.5, & \text{hard mixed group},\\
1.0, & \text{otherwise}.
\end{cases}
\]
A hard mixed group has at least one successful candidate, but no successful candidate among the
baseline local offsets $\delta\in\{-1,0,+1\}$. These are cases where recovery requires selecting a
farther or less obvious milestone.

\paragraph{Actor update.}
The selector scores define a categorical distribution over candidates:
\[
    \pi_\phi(c\mid o_f,H_f,C_f)
    =
    \mathrm{softmax}(s)_c.
\]
We update the selector with
\begin{equation}
\label{eq:app_lactor}
    L_{\mathrm{actor}}
    =
    \mathbb{E}
    \left[
        w_f
        \sum_{c\in C_f}
        \pi_\phi(c\mid o_f,H_f,C_f)
        \left(
            \alpha
            \log \pi_\phi(c\mid o_f,H_f,C_f)
            -
            \min_{k\in\{1,2\}}
            Q_k(c\mid o_f,H_f,C_f)
        \right)
    \right].
\end{equation}
The lower-entropy actor term encourages a sharper distribution over the small local candidate set.
We initialize the temperature to 0.05 and learn it during fine-tuning.

\paragraph{Positive-candidate regularization.}
To preserve the supervised preference for observed successful candidates, we reuse the grouped
positive-candidate objective from the supervised warm-start as a behavior-cloning regularizer. For
groups with at least one successful candidate, let
\[
    q_c =
    \frac{y_c}{\sum_{c'\in C_f} y_{c'}} .
\]
The corresponding positive-candidate BC loss is
\[
    L_{\mathrm{BC}}
    =
    -
    \sum_{c\in C_f}
    q_c
    \log
    \pi_\phi(c\mid o_f,H_f,C_f).
\]
Since $\pi_\phi(c\mid o_f,H_f,C_f)=\mathrm{softmax}(s)_c$, this is equivalent to the grouped
positive-candidate ranking objective:
\[
    L_{\mathrm{BC}}
    =
    -
    \frac{1}{Y_f}
    \sum_{c\in C_f}
    y_c s_c
    +
    \log
    \sum_{c\in C_f}
    \exp(s_c)
    =
    L_{\mathrm{group}},
    \qquad
    Y_f=\sum_{c\in C_f}y_c .
\]
For all-fail groups, this term is omitted. The final fine-tuning objective can therefore be written as
\[
    L_{\mathrm{ft}}
    =
    L_Q
    +
    L_{\mathrm{actor}}
    +
    \lambda_{\mathrm{BC}} L_{\mathrm{BC}},
    \qquad
    \lambda_{\mathrm{BC}}=0.5.
\]

\paragraph{Optimization.}
We fine-tune for 8 epochs with 200 optimization steps per epoch. The main selector parameters use
AdamW with learning rate $10^{-4}$, while the pair projector uses a reduced learning rate
$3\times10^{-5}$, corresponding to the $0.3\times$ multiplier described in
Sec.~\ref{app:step_pair_projectors}. We use weight decay $10^{-4}$, batch size 16, and gradient
clipping at 1.0. The selected fine-tuned checkpoint is the one with the best validation top-1 recovery
success.

\subsection{Selector Cost and Test-Time Overhead}
\label{app:selector_cost_overhead}

B2FF separates three costs: pre-execution familiar-bank construction, offline counterfactual rollout
collection, and post-trigger test-time recovery. Counterfactual candidate rollouts are used only for
offline selector training and candidate-oracle analysis. They are never executed during test-time
recovery.

\begin{table}[H]
\centering
\caption{
Compute summary for B2FF. Pre-execution bank construction is paid before robot action execution.
Post-trigger quantities are measured after the familiar future bank has already been constructed.
}
\label{tab:b2ff_compute_summary}
\footnotesize
\setlength{\tabcolsep}{4.5pt}
\renewcommand{\arraystretch}{1.06}
\begin{tabularx}{\linewidth}{@{}p{0.34\linewidth}p{0.18\linewidth}X@{}}
\toprule
Component & Mean latency & Interpretation \\
\midrule
Pre-execution bank construction
& 190.2 s
& End-to-end startup time from run start until the familiar future bank is available. \\

\midrule
Nominal VLA call
& 7.47 s
& Standard joint future--action generation. \\

Failed-state re-plan call
& 7.57 s
& Re-predicts a future from the failed observation before action generation. \\

Fixed-milestone recovery call
& 2.15 s
& Action-only denoising with a fixed familiar milestone. \\

Selector-side overhead
& 7.27 ms
& One selector forward pass plus local candidate construction at recovery entry. \\
\bottomrule
\end{tabularx}
\end{table}

The pre-execution bank-construction cost is not included in post-trigger recovery latency because
the bank is generated before any failure is detected. Given a pre-built bank, the selector-side
overhead is incurred once per recovery trigger, not once per recovery chunk. In contrast, the frozen
VLA action-generation cost scales with the number of recovery chunks: for a recovery window of
$W$ chunks, B2FF can issue up to $W$ fixed-milestone action-only VLA calls. In the main protocol,
$W=5$, and the selected milestone is kept fixed throughout this window.

The latency comparison therefore separates two effects. Failed-state re-planning performs future
generation from the off-trajectory observation and takes 7.57 s on average for the recovery call.
B2FF instead reuses a pre-imagined milestone and performs action-only denoising, which takes
2.15 s per recovery action-generation call. The selector-side overhead is negligible compared with
either VLA call.

The familiar future bank is lightweight to store. Each milestone is represented by 725 visual token
ids, corresponding to a $25\times25$ full-view grid and a $10\times10$ wrist-view grid. For an
initial bank of $M_0=12$ milestones, the bank requires about 68 KiB per episode when stored as
int64 token ids.
\section{Failure-Injected LIBERO Benchmark}
\label{app:failure_injected_libero}

\label{app:benchmark_construction}
\label{app:failure_type_definitions}
\label{app:perturbation_ranges}

Failure-injected LIBERO evaluates whether a policy can recover from a bounded off-trajectory
deviation while the task remains physically feasible. Each episode starts from the standard LIBERO
task and initial-state distribution. We execute the frozen VLA under nominal control, inject one
perturbation at a sampled failure time $t_{\mathrm{fail}}$, and then evaluate recovery from the
resulting off-trajectory state. All methods are evaluated on the same held-out perturbation
instances: the task, initial state, language instruction, failure type, perturbation parameters, and
injection time are fixed across methods.

We use three recoverable perturbation families. \textit{Gripper XY offset} models end-effector
misalignment by applying a bounded random XY offset to the gripper. \textit{Object XY shift}
models target-object displacement by shifting the target or goal object's MuJoCo joint pose in the
XY plane. \textit{Object laydown} models an unexpected object pose change by applying an X-axis
or Y-axis tilt to the object's free-joint quaternion, with a small lift when needed to avoid invalid
contact. All perturbations are bounded so that the relevant object remains observable, reachable,
and inside the workspace.

\begin{table}[H]
\centering
\caption{
Failure-injection magnitude ranges. Each failure-injected episode contains one perturbation sampled
from the corresponding range.
}
\label{tab:failure_injection_specs}
\small
\setlength{\tabcolsep}{6pt}
\renewcommand{\arraystretch}{1.08}
\begin{tabular}{@{}ll@{}}
\toprule
Failure type & Sampled magnitude \\
\midrule
Gripper XY offset & 4--6 / 6--8 / 8--10 cm \\
Object XY shift & 1--2 / 3--4 / 5--6 cm \\
Object laydown & 70--85$^\circ$ or 85--100$^\circ$ tilt \\
\bottomrule
\end{tabular}
\end{table}

The offline counterfactual recovery groups used to train the selector are generated with the same
failure-injection procedure, but are separate from the held-out evaluation episodes. Their
candidate-wise rollout labels are used only for selector training and candidate-oracle analysis, not
during test-time recovery.

\section{Baselines and Selection Rules}
\label{app:baselines_selection_rules}
\label{app:shared_recovery_protocol}
\label{app:selection_rules}
\label{app:selection_rule_summary}

For B2FF-specific selection-rule comparisons, all deployable recovery methods use the same frozen
VLA, familiar future bank, recovery entry index, candidate offset set, and recovery window. They
differ only in the recovery-time visual condition. No deployable baseline updates the VLA weights
or performs candidate-wise trial rollouts during test-time recovery. The candidate oracle is reported
only as an offline upper bound over the same local candidate set.

\begin{table}[H]
\centering
\caption{
Summary of recovery-time selection rules. Deployable methods use the same frozen VLA and differ
only in the visual condition used during recovery.
}
\label{tab:selection_rule_summary}
\footnotesize
\setlength{\tabcolsep}{4pt}
\renewcommand{\arraystretch}{1.05}
\begin{tabularx}{\linewidth}{@{}p{0.24\linewidth}Xcc@{}}
\toprule
Method & Recovery visual condition & Uses bank? & Deployable? \\
\midrule
UD-VLA / no-fix
& Nominal VLA execution without explicit recovery
& No & Yes \\

Failed-state re-plan
& Future image re-predicted once from the failed observation $o_f$
& No & Yes \\

Fixed-Previous
& Familiar milestone $\tilde v_{f-1}$
& Yes & Yes \\

Fixed-Current
& Familiar milestone $\tilde v_f$
& Yes & Yes \\

Fixed-Near
& Familiar milestone $\tilde v_{f+1}$
& Yes & Yes \\

Random candidate
& Uniformly sampled valid candidate from $C_f$
& Yes & Yes \\

Visual-nearest
& Candidate in $C_f$ most similar to $o_f$ in frozen visual-tokenizer feature space
& Yes & Yes \\

B2FF
& Candidate in $C_f$ with the highest learned recoverability score
& Yes & Yes \\

Candidate oracle
& Any candidate in $C_f$ with a successful counterfactual rollout label
& Yes & No \\
\bottomrule
\end{tabularx}
\end{table}

\noindent
\textsc{Best-Fixed}, used in the main-text visualization, denotes the strongest aggregate result among
\textsc{Fixed-Previous}, \textsc{Fixed-Current}, and \textsc{Fixed-Near}. It is not a per-episode
oracle and is not treated as a separate deployable method.
\section{Protocol Ablations}
\label{app:protocol_ablations}

This section reports compact diagnostics for two protocol choices: recovery-window length and
trigger timing. Candidate-set composition is reported in Sec.~\ref{app:selector_arch_ablation}.

\subsection{Recovery Window Length}
\label{app:recovery_window_ablation}

The recovery window $W$ is the number of consecutive decision chunks for which the selected
familiar milestone is kept fixed. Table~\ref{tab:recovery_window_ablation} reports representative
window lengths. Performance changes only modestly across the tested values, so we treat $W$ as a
fixed protocol parameter rather than a method-specific tuning knob. The final protocol uses $W=5$.

\begin{table}[H]
\centering
\caption{
Recovery-window ablation. Values are average task success rates (\%) across the four
failure-injected LIBERO suites. The $W=5$ column is the final setting used in the main B2FF
protocol.
}
\label{tab:recovery_window_ablation}
\footnotesize
\setlength{\tabcolsep}{8pt}
\renewcommand{\arraystretch}{1.05}
\begin{tabular}{@{}lccc@{}}
\toprule
Metric & $W=2$ & $W=4$ & $W=5$ \\
\midrule
Average success & 73.5 & 73.8 & \textbf{74.0} \\
\bottomrule
\end{tabular}
\end{table}

The small variation across window lengths suggests that B2FF's gains are driven primarily by
recoverability-aware milestone selection rather than by finely tuning the recovery duration. We use
$W=5$ because it remains competitive while providing a slightly longer fixed visual anchor for
returning toward a familiar trajectory.

\subsection{Trigger Timing}
\label{app:trigger_timing_ablation}

We study sensitivity to delayed recovery entry by shifting the recovery trigger later than the
controlled perturbation-aligned entry point. The $+0$ setting corresponds to controlled recovery
timing. All settings use the same familiar future bank, candidate construction procedure, selector,
and recovery window.

\begin{table}[H]
\centering
\caption{
Trigger-timing ablation on failure-injected LIBERO-Goal. The trigger delay is measured in
environment steps relative to the controlled perturbation-aligned recovery entry. Values are task
success rates (\%), rounded to one decimal.
}
\label{tab:trigger_timing_ablation}
\footnotesize
\setlength{\tabcolsep}{8pt}
\renewcommand{\arraystretch}{1.05}
\begin{tabular}{@{}lc@{}}
\toprule
Trigger delay & Success (\%) \\
\midrule
$+0$  & \textbf{73.3} \\
$+10$ & 65.8 \\
$+20$ & 60.0 \\
$+30$ & 61.7 \\
$+40$ & 58.3 \\
\bottomrule
\end{tabular}
\end{table}

Delayed recovery generally reduces success. As the delay grows, the perturbation can move the
rollout farther from the familiar trajectory and leave less margin for the selected milestone to guide
the frozen VLA back to a recoverable state. This supports using perturbation-aligned timing in the
controlled comparison to isolate milestone selection, while stronger online trigger estimation remains
an important direction for future work.
\definecolor{B2FFBlue}{HTML}{2F80ED}
\definecolor{B2FFBg}{HTML}{EAF3FF}
\definecolor{OracleGold}{HTML}{D4A017}
\definecolor{OracleBg}{HTML}{FFF4D6}

\section{Additional Simulation Results}
\label{app:additional_sim_results}

\subsection{Controlled Selection-Rule Results by Suite and Failure Type}
\label{app:selection_rule_ablation_by_suite}

Table~\ref{tab:selection_rule_ablation_per_suite} expands the controlled selection-rule comparison
by LIBERO suite and injected failure type. All bank-based recovery methods use the same frozen
VLA, familiar future bank, recovery entry index, candidate set, and recovery window. They differ
only in which visual condition is fixed during recovery. The candidate oracle is an offline upper
bound over the same local candidate set and is not a deployable test-time method.

\begin{table*}[t]
\centering
\scriptsize
\setlength{\tabcolsep}{2.8pt}
\renewcommand{\arraystretch}{1.12}

\resizebox{\textwidth}{!}{%
\begin{tabular}{@{}lrrrrrrrrrrrrrrrr@{}}
\toprule
& \multicolumn{4}{c}{\textbf{LIBERO-Object}}
& \multicolumn{4}{c}{\textbf{LIBERO-Spatial}}
& \multicolumn{4}{c}{\textbf{LIBERO-Goal}}
& \multicolumn{4}{c}{\textbf{LIBERO-Long}} \\
\cmidrule(lr){2-5}
\cmidrule(lr){6-9}
\cmidrule(lr){10-13}
\cmidrule(lr){14-17}
\textbf{Selection rule}
& \textbf{All} & \textbf{Grip.} & \textbf{Obj.} & \textbf{Lay.}
& \textbf{All} & \textbf{Grip.} & \textbf{Obj.} & \textbf{Lay.}
& \textbf{All} & \textbf{Grip.} & \textbf{Obj.} & \textbf{Lay.}
& \textbf{All} & \textbf{Grip.} & \textbf{Obj.} & \textbf{Lay.} \\
\midrule

\multicolumn{17}{@{}l}{\textsc{Nominal no-recovery baseline}} \\
UD-VLA / no-fix
& \heat{52.5} & \heat{37.5} & \heat{52.5} & \heat{67.5}
& \heat{58.3} & \heat{75.0} & \heat{52.5} & \heat{47.5}
& \heat{58.3} & \heat{47.5} & \heat{60.0} & \heat{67.5}
& \heat{55.8} & \heat{62.5} & \heat{65.0} & \heat{40.0} \\
\midrule

\multicolumn{17}{@{}l}{\textsc{Heuristic recovery rules}} \\
Failed-state re-plan
& \heat{54.2} & \heat{47.5} & \heat{62.5} & \heat{52.5}
& \heat{51.7} & \heat{70.0} & \heat{52.5} & \heat{32.5}
& \heat{63.3} & \heat{62.5} & \heat{60.0} & \heat{67.5}
& \heat{69.2} & \heat{80.0} & \heat{75.0} & \heat{52.5} \\

Fixed previous
& \heat{50.8} & \heat{50.0} & \heat{55.0} & \heat{47.5}
& \heat{55.0} & \heat{82.5} & \heat{50.0} & \heat{32.5}
& \heat{60.0} & \heat{62.5} & \heat{57.5} & \heat{60.0}
& \heat{70.0} & \heat{70.0} & \heat{72.5} & \heat{67.5} \\

Fixed current
& \heat{57.5} & \heat{52.5} & \heat{52.5} & \heat{67.5}
& \heat{50.0} & \heat{70.0} & \heat{52.5} & \heat{27.5}
& \heat{58.3} & \heat{60.0} & \heat{55.0} & \heat{60.0}
& \heat{58.3} & \heat{57.5} & \heat{70.0} & \heat{47.5} \\

Fixed near
& \heat{30.8} & \heat{27.5} & \heat{37.5} & \heat{27.5}
& \heat{50.0} & \heat{75.0} & \heat{55.0} & \heat{20.0}
& \heat{55.0} & \heat{65.0} & \heat{60.0} & \heat{40.0}
& \heat{66.7} & \heat{85.0} & \heat{62.5} & \heat{52.5} \\

Random candidate
& \heat{49.2} & \heat{47.5} & \heat{50.0} & \heat{50.0}
& \heat{52.5} & \heat{77.5} & \heat{52.5} & \heat{27.5}
& \heat{59.2} & \heat{67.5} & \heat{60.0} & \heat{50.0}
& \heat{67.5} & \heat{75.0} & \heat{67.5} & \heat{60.0} \\

Visual-nearest
& \heat{50.8} & \heat{47.5} & \heat{52.5} & \heat{52.5}
& \heat{53.3} & \heat{75.0} & \heat{52.5} & \heat{32.5}
& \heat{59.2} & \heat{70.0} & \heat{60.0} & \heat{47.5}
& \heat{61.7} & \heat{72.5} & \heat{65.0} & \heat{47.5} \\

\arrayrulecolor{B2FFBlue}
\midrule
\rowcolor{B2FFBg}
\textbf{B2FF}
& \bhe{69.2} & \bhe{67.5} & \bhe{72.5} & \bhe{67.5}
& \bhe{65.8} & \bhe{92.5} & \bhe{65.0} & \bhe{40.0}
& \bhe{73.3} & \bhe{82.5} & \bhe{70.0} & \bhe{67.5}
& \bhe{87.5} & \bhe{100.0} & \bhe{87.5} & \bhe{75.0} \\
\arrayrulecolor{black}
\midrule

\multicolumn{17}{@{}l}{\textsc{Offline upper bound}} \\
\rowcolor{OracleBg}
Candidate oracle
& \bocell{84.2} & \bocell{87.5} & \bocell{85.0} & \bocell{80.0}
& \bocell{69.2} & \bocell{95.0} & \bocell{65.0} & \bocell{47.5}
& \bocell{75.8} & \bocell{82.5} & \bocell{72.5} & \bocell{72.5}
& \bocell{92.5} & \bocell{100.0} & \bocell{90.0} & \bocell{87.5} \\

\bottomrule
\end{tabular}
}

\caption{
Controlled familiar-future selection-rule ablation by LIBERO suite and injected failure type.
Darker cells indicate higher success. The \textit{All} columns follow the same suite-level aggregation
used in the main comparison, rounded to one decimal. The \textit{Grip.}, \textit{Obj.}, and
\textit{Lay.} columns provide diagnostic breakdowns by injected failure type. All bank-based
methods use the same frozen VLA, familiar future bank, recovery entry index, candidate set, and
recovery window, and differ only in the recovery visual condition selected after failure. The candidate
oracle is an offline upper bound over the same local candidate set and is not a deployable test-time
method.
}
\label{tab:selection_rule_ablation_per_suite}
\end{table*}

\subsection{Selector Ranking Diagnostic}
\label{app:selector_rank_diagnostic}

Table~\ref{tab:selector_rank_diagnostic} analyzes the ranking produced by the learned selector.
For each recovery context, we sort the valid candidate milestones by selector score and evaluate the
counterfactual success rate of the candidate at each rank. Rank 1 corresponds to the candidate that
B2FF would select at test time.

\begin{table}[H]
\centering
\caption{
Selector rank diagnostic on failure-injected LIBERO. Candidate milestones are sorted by the learned
selector score within each recovery context. Values are counterfactual task success rates (\%) for
the candidate at each rank. 
}
\label{tab:selector_rank_diagnostic}
\footnotesize
\setlength{\tabcolsep}{6pt}
\renewcommand{\arraystretch}{1.05}
\begin{tabular}{@{}lccccc@{}}
\toprule
Suite & Rank 1 & Rank 2 & Rank 3 & Rank 4 & Rank 5 \\
\midrule
LIBERO-Object  & 69.2 & 50.0 & 48.3 & 41.7 & 37.5 \\
LIBERO-Spatial & 65.8 & 58.3 & 55.0 & 59.2 & 60.0 \\
LIBERO-Goal    & 73.3 & 61.7 & 64.2 & 61.7 & 63.3 \\
LIBERO-Long    & 87.5 & 66.7 & 62.5 & 65.0 & 65.0 \\
\midrule
Average        & \textbf{74.0} & 59.2 & 57.5 & 56.9 & 56.5 \\
\bottomrule
\end{tabular}
\end{table}

The rank diagnostic shows that the learned milestone selector produces a meaningful
recoverability ordering over the local candidate set. The top-ranked candidate, which is the
candidate selected by B2FF at test time, achieves the highest average counterfactual success rate
across suites. Lower-ranked candidates are weaker on average, indicating that the selector score is
not merely separating valid from invalid bank entries but is learning to prioritize milestones that
are more likely to support successful recovery from the current failure context.
\begin{figure}[p]
\centering
\includegraphics[width=0.96\linewidth,height=0.84\textheight,keepaspectratio]{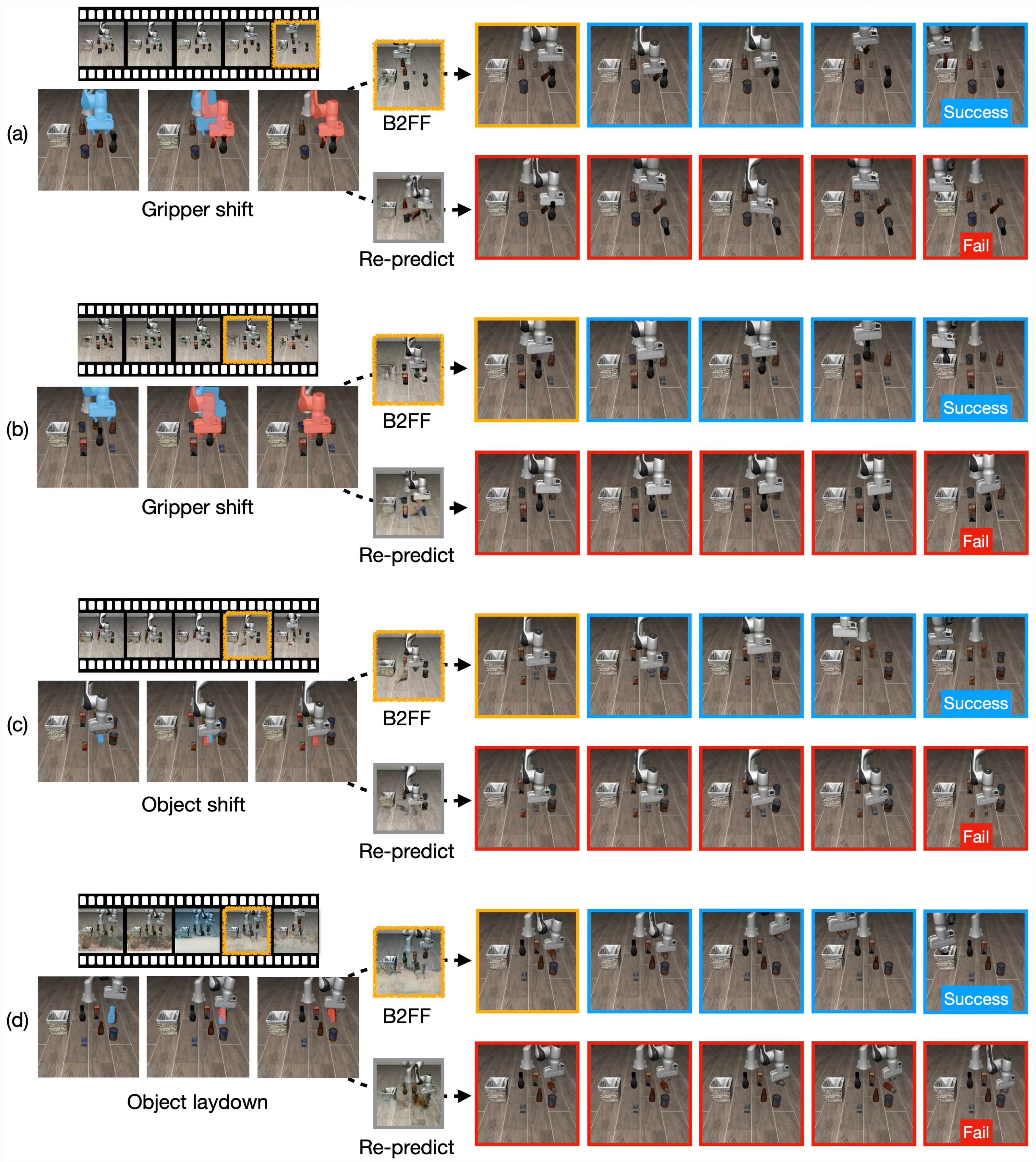}
\caption{
Additional qualitative simulation recovery cases. Rows show \textbf{(a--b)} gripper-shift
perturbations, \textbf{(c)} object-shift perturbation, and \textbf{(d)} object-laydown perturbation.
In each row, the filmstrip shows the familiar future bank generated before execution, and the yellow
box marks the milestone selected by B2FF. Starting from the same perturbed observation, B2FF
clamps the selected pre-imagined milestone as the recovery condition and reaches task success,
whereas failed-state re-prediction (\emph{Re-predict}) generates a new future from the
off-trajectory observation and fails. Blue and red borders denote successful and failed rollouts,
respectively.
}
\label{fig:sim_qualitative_recovery}
\end{figure}

\subsection{Additional Qualitative Simulation Recovery Cases}
\label{app:sim_qualitative_cases}

Figure~\ref{fig:sim_qualitative_recovery} provides additional qualitative simulation examples under
controlled failure injection. Each row starts from a perturbed observation and compares B2FF against
failed-state re-prediction. B2FF selects a milestone from the pre-execution familiar future bank and
uses it as the fixed future-image condition for action-only denoising. In contrast, failed-state
re-prediction generates a new future directly from the off-trajectory observation.

Across gripper-shift, object-shift, and object-laydown perturbations, these examples illustrate that
B2FF can recover by anchoring the frozen VLA to a familiar, pre-imagined milestone, whereas
re-predicting from the failed observation often produces an unstable recovery condition. This
qualitative pattern is consistent with the quantitative selection-rule results: recovery depends not
only on having a familiar future bank, but also on selecting a milestone that is recoverable from the
current off-trajectory state.

\section{Real-World Experiments}
\label{app:real_world_experiments}

This section provides additional details for the real-world recovery experiments summarized in the
main text. We evaluate B2FF on a tabletop robot setup across three manipulation tasks and two
recoverable injected failure types. The real-world VLA is trained for nominal task execution and
then kept frozen; B2FF only tunes the lightweight milestone selector for real-world recovery.

\subsection{Robot Setup and Tasks}
\label{app:real_world_robot_setup}

We use a tabletop manipulation setup with a 6-DoF Universal Robots UR5 arm equipped with a
two-finger gripper. All episodes begin from a standard home pose, and objects are placed within the
tabletop workspace so that they remain reachable by the robot during both nominal execution and
recovery trials.

The real-world setup uses two camera views. The main third-person camera is an Azure Kinect DK,
mounted at a fixed position to observe the full tabletop scene. The wrist-view camera is The
RealSense\texttrademark{} Depth Camera D405, mounted near the robot wrist to provide a close-up
view of the gripper-object interaction region. Both camera poses are fixed throughout data
collection and evaluation.

\begin{figure}[H]
\centering
\includegraphics[width=0.95\linewidth]{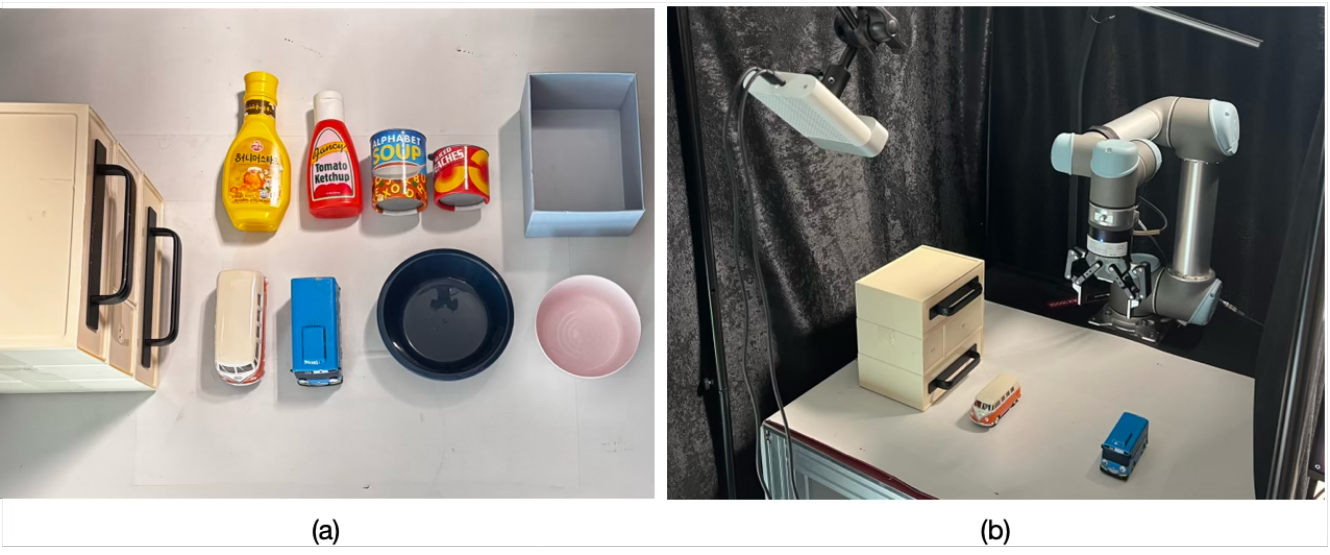}
\caption{
\textbf{(a)} Objects used across the real-world tasks.
\textbf{(b)} Tabletop robot setup.
}
\label{fig:real_world_setup_objects}
\end{figure}

Figure~\ref{fig:real_world_setup_objects} shows the physical setup and object set. We evaluate
three manipulation tasks: object stacking, pick-and-place, and drawer closing with object placement.
The object set includes everyday tabletop items with different shapes, sizes, and contact properties,
such as toy vehicles, bottles, cans, bowls, and containers.

\subsection{Real-World VLA and Selector Tuning}
\label{app:real_world_selector_tuning}

The underlying real-world VLA policy is trained from 600 nominal demonstration trajectories
collected across the three tasks. 

For real-world recovery, we initialize the milestone selector from the LIBERO-trained selector
checkpoint and tune only the selector on 35 real-world recovery groups. Each recovery group is
constructed by creating a shared failure context, building a familiar future bank, forming a local
candidate set, and evaluating candidate milestones offline by clamping each candidate as the
recovery visual condition. These candidate-wise labels are used only for selector tuning, not for
test-time recovery.

Table~\ref{tab:real_world_config_tuning} summarizes the real-world details needed to reproduce
the recovery setup. We omit low-level VLA decoding hyperparameters that are unchanged from the
frozen nominal policy, since B2FF does not tune or modify the VLA backbone. The relevant
execution unit for recovery is the decision chunk: in the real-world setup, each decision chunk
executes 8 low-level actions.

\begin{table}[H]
\centering
\caption{
Real-world VLA and selector tuning summary. The real-world VLA is trained for nominal execution
and kept frozen; only the milestone selector is tuned for recovery.
}
\label{tab:real_world_config_tuning}
\footnotesize
\setlength{\tabcolsep}{5pt}
\renewcommand{\arraystretch}{1.08}
\begin{tabularx}{\linewidth}{@{}p{0.33\linewidth}X@{}}
\toprule
Item & Value \\
\midrule
Nominal VLA data
& 600 nominal demonstrations across three tasks \\

Observations
& Static $256\times256$ and wrist $80\times80$ images \\

Execution unit
& 8 low-level actions per decision chunk \\

Frozen components
& Visual tokenizer, future-image generator, action denoiser, and VLA backbone;
0 VLA backbone updates \\

Selector adaptation data
& 35 real-world recovery groups; 3 candidates/group with offsets $\{-1,0,+1\}$;
105 offline candidate-wise labels \\

Selector tuning
& Initialized from the LIBERO-trained selector; 8 epochs, 60 steps/epoch,
batch size 8, learning rate $5\times10^{-5}$ \\

Test-time recovery
& $W_{\mathrm{real}}=1$ decision chunk; one selector forward pass per recovery trigger;
0 test-time candidate-wise trial rollouts \\
\bottomrule
\end{tabularx}
\end{table}


\subsection{Real-World Failure Injection}
\label{app:real_world_failure_injection}

We evaluate two recoverable real-world injected failure types: gripper shift and target-object shift.
Both perturbations are bounded so that the relevant object remains visible, inside the tabletop
workspace, and reachable by the robot.

\paragraph{Gripper shift.}
This perturbation displaces the end-effector away from its nominal interaction pose. It models
misalignment during grasping or contact while preserving the feasibility of the underlying task.

\paragraph{Target-object shift.}
This perturbation displaces the target object from its nominal trajectory while keeping it inside the
workspace and visible to the policy cameras. It models unexpected object motion or contact-induced
displacement.

\begin{figure}[t]
\centering
\includegraphics[width=0.86\linewidth]{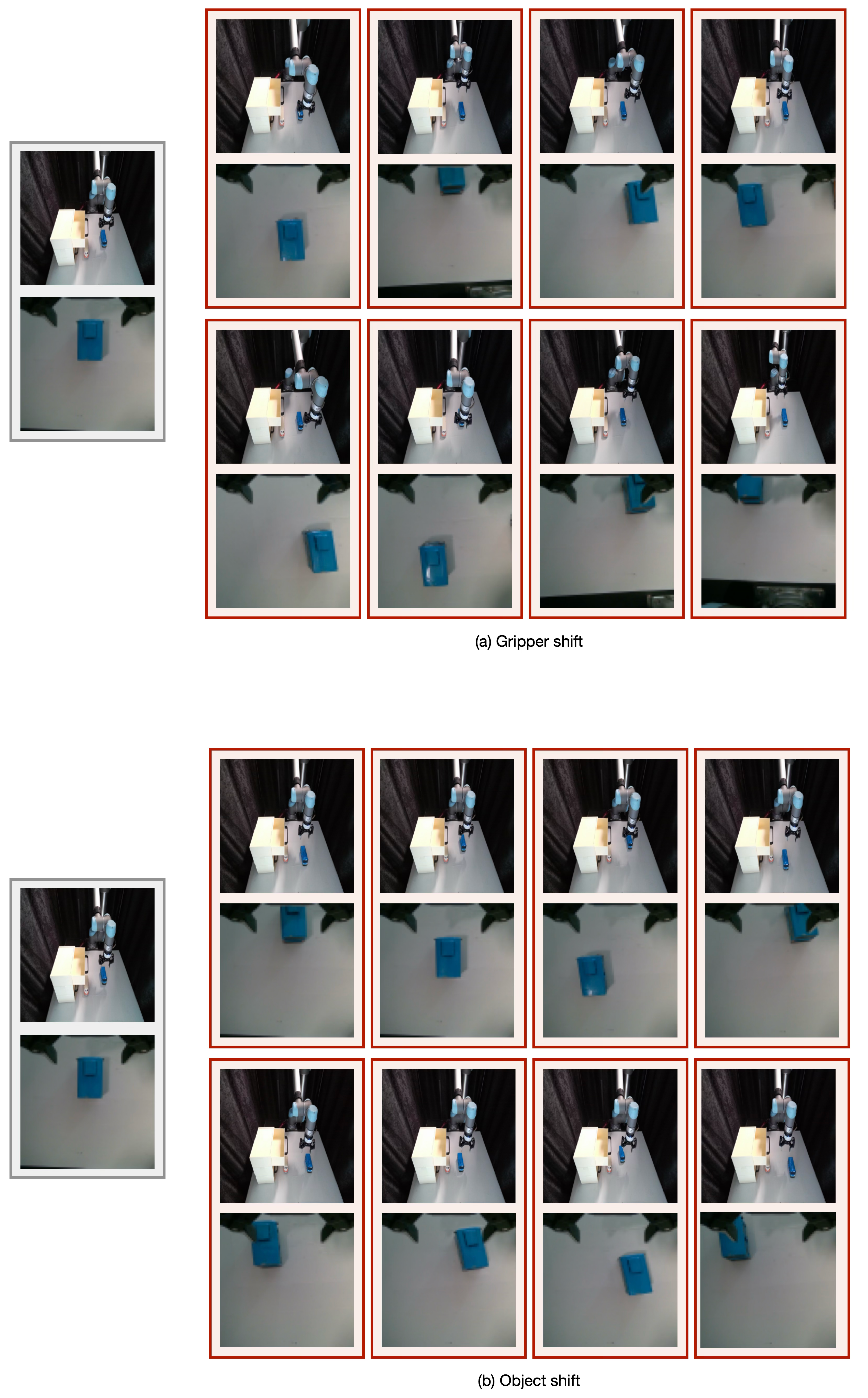}
\caption{
Real-world perturbation visualization. \textbf{(a)} Gripper shift displaces the end-effector from its
nominal interaction pose while keeping the target object reachable. \textbf{(b)} Target-object shift
moves the target object within the tabletop workspace while preserving visibility and reachability.
For each failure type, the left block shows the paired static and wrist observations around the
perturbation context, and the right block shows representative post-perturbation observations across
real-world trials. 
}
\label{fig:real_world_perturbation_visualization}
\end{figure}

Figure~\ref{fig:real_world_perturbation_visualization} visualizes representative perturbation
contexts for both failure types. These examples illustrate the injected deviations rather than
recovery outcomes.

We exclude trials in which the relevant object leaves the reachable workspace, becomes severely
occluded, or enters a state that cannot be recovered by the underlying manipulation skill.

\subsection{Real-World Transfer and Tuning Result}
\label{app:real_world_breakdown}

The real-world evaluation contains 90 recovery trials across the three tasks and two injected failure
types. Table~\ref{tab:real_world_zero_shot_tuning} reports the overall transfer and tuning result.
The zero-shot selector is initialized from the LIBERO-trained checkpoint and evaluated without
real-world selector tuning. Fine-tuning the selector on 35 real-world recovery groups improves the
overall success rate from 55.6\% to 61.1\%.

\begin{table}[H]
\centering
\caption{
Real-world selector transfer and tuning. Values are success rates with success counts in
parentheses over 90 real-world recovery trials. The zero-shot selector uses the LIBERO-trained
checkpoint without real-world recovery-group tuning.
}
\label{tab:real_world_zero_shot_tuning}
\footnotesize
\setlength{\tabcolsep}{5pt}
\renewcommand{\arraystretch}{1.05}
\begin{tabular}{@{}lcc@{}}
\toprule
Selector variant & Real-world tuning groups & Overall success \\
\midrule
Zero-shot B2FF selector & 0  & 55.6\% (50/90) \\
Real-world tuned B2FF selector & 35 & 61.1\% (55/90) \\
\bottomrule
\end{tabular}
\end{table}

These results indicate that the LIBERO-trained selector transfers nontrivially to the real-world
domain, while a small amount of real-world recovery-group tuning improves the selector's
recoverability ranking under the physical robot setup.

\subsection{Additional Real-World Qualitative Recovery Cases}
\label{app:real_world_qualitative_cases}

\begin{figure}[p]
\centering
\includegraphics[width=0.90\linewidth,height=0.88\textheight,keepaspectratio]{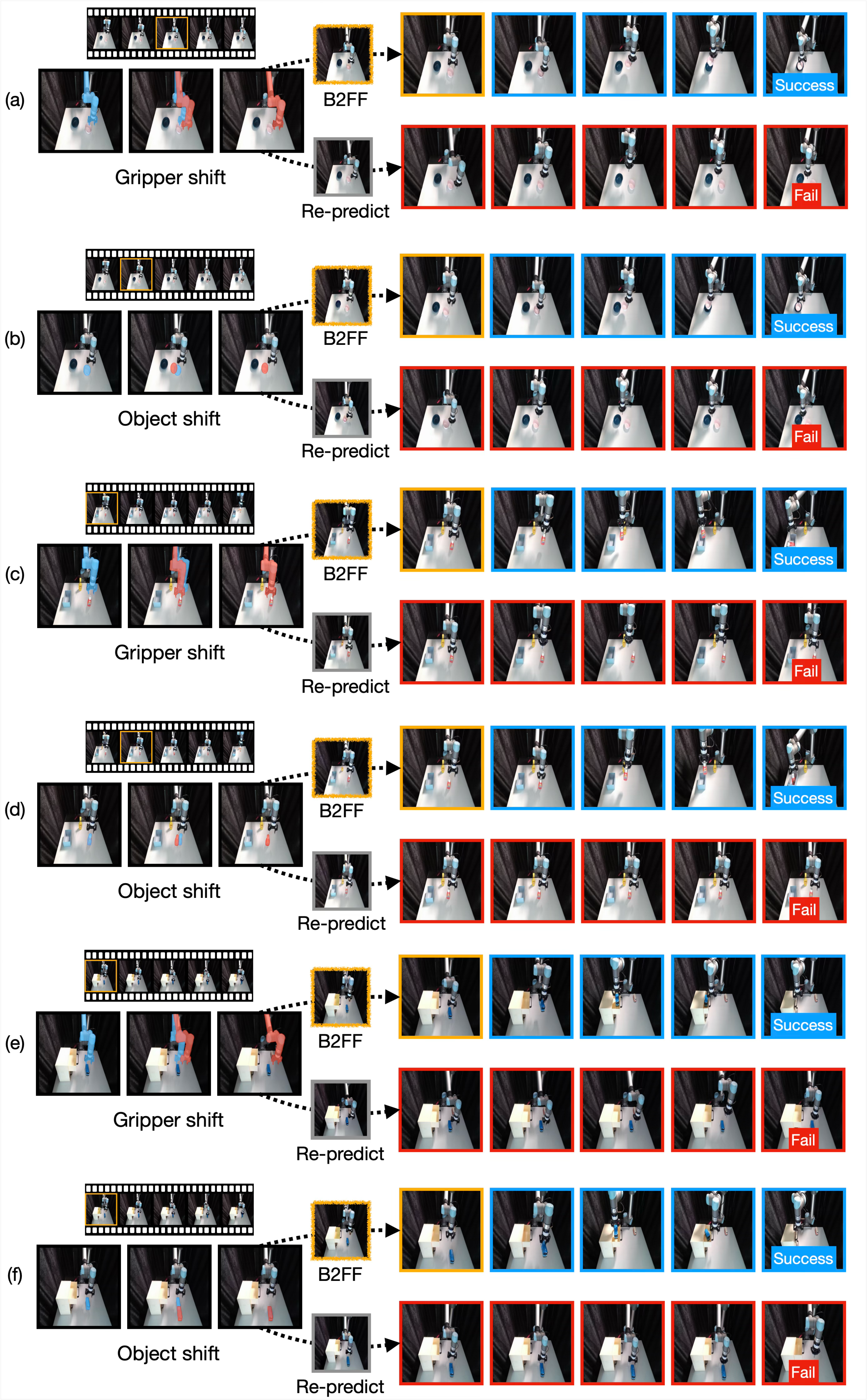}
\caption{
Additional real-world qualitative recovery cases. Rows are grouped by task: \textbf{(a--b)} Task 1,
\textbf{(c--d)} Task 2, and \textbf{(e--f)} Task 3. Within each task pair, the first row shows a
gripper-shift perturbation and the second row shows a target-object-shift perturbation. In each row,
the filmstrip shows the familiar future bank generated before execution, and the yellow box marks
the milestone selected by B2FF. Starting from the same perturbed observation, B2FF clamps the
selected pre-imagined milestone as the recovery condition and reaches task success, whereas
failed-state re-prediction (\emph{Re-predict}) generates a new future from the off-trajectory
observation and fails. Blue and red borders denote successful and failed rollouts, respectively.
}
\label{fig:real_world_qualitative_recovery}
\end{figure}

Figure~\ref{fig:real_world_qualitative_recovery} shows additional qualitative recovery examples on
the real robot. Rows are grouped by task: the first two rows correspond to Task 1, the next two rows
to Task 2, and the last two rows to Task 3. Within each task pair, the first row shows a gripper-shift
perturbation and the second row shows a target-object-shift perturbation.

\end{document}